\pdfoutput=1

\documentclass[11pt]{article}

\usepackage{acl}

\usepackage{times}
\usepackage{latexsym}

\usepackage[T1]{fontenc}

\usepackage[utf8]{inputenc}

\usepackage{microtype}

\usepackage{inconsolata}

\usepackage{amsmath}
\usepackage{booktabs}
\usepackage{color}
\usepackage{graphicx}
\usepackage{makecell}
\usepackage{mathtools}
\usepackage{multirow}
\usepackage{paralist}
\usepackage{xltabular}
\usepackage{xspace}

\newcommand{\modelname}{MAGIC\xspace}
\newcommand{\addnew}[1]{{\color{black}{#1}}}
\newcommand{\green}[1]{{\color{blue}{#1}}}

\title{Multi-Aspect Controllable Text Generation with Disentangled Counterfactual Augmentation}

\author{ 
    Yi Liu$^\dagger$ \quad
    Xiangyu Liu$^\dagger$ \quad
    Xiangrong Zhu$^\dagger$ \quad
    Wei Hu$^{\dagger,\,\ddagger,\,}$\thanks{\,\, Corresponding author} \\
    $^\dagger$ State Key Laboratory for Novel Software Technology, Nanjing University, China \\
    $^\ddagger$ National Institute of Healthcare Data Science, Nanjing University, China \\
    \texttt{\{yiliu07, xyl, xrzhu\}.nju@gmail.com, whu@nju.edu.cn}}

\begin{document}
\maketitle

\begin{abstract}
Multi-aspect controllable text generation aims to control the generated texts in attributes from multiple aspects (e.g., ``positive'' from \textit{sentiment} and ``sport'' from \textit{topic}). 
For ease of obtaining training samples, existing works neglect attribute correlations formed by the intertwining of different attributes. 
Particularly, the stereotype formed by imbalanced attribute correlations significantly affects multi-aspect control.
In this paper, we propose \modelname, a new \textbf{m}ulti-\textbf{a}spect controllable text \textbf{g}eneration method with d\textbf{i}sentangled \textbf{c}ounterfactual augmentation.
We alleviate the issue of imbalanced attribute correlations during training using counterfactual feature vectors in the attribute latent space by disentanglement.
During inference, we enhance attribute correlations by target-guided counterfactual augmentation to further improve multi-aspect control.
Experiments show that \modelname outperforms state-of-the-art baselines in both imbalanced and balanced attribute correlation scenarios.
Our source code and data are available at \url{https://github.com/nju-websoft/MAGIC}.
\end{abstract}

\section{Introduction}
Controllable text generation (CTG) aims to generate texts adhering to given constraints reliably.
The development of generative AI based on large language models (LLMs) draws increasing attention to CTG \citep{ctrl2019Keskar, gpt2020Brown, opt2022Zhang}.
Due to the demand for diverse attribute control, recent studies focus on a more practical and challenging setting, \emph{multi-aspect controllable text generation} (MCTG).
Different kinds of methods have been proposed\citep{discrete2022Gu}, including weighted decoding \citep{PPLM2020Dath, GEDI2021Krause}, optimization in the language space \citep{mucoco2021Kumar, mix_match2022Mire}, optimization in the latent semantic space \citep{discrete2022Gu, prior2023Gu, Mac2023Ding, OED2023Liu}, etc. 

\begin{figure}
\centering
\includegraphics[width=\columnwidth]{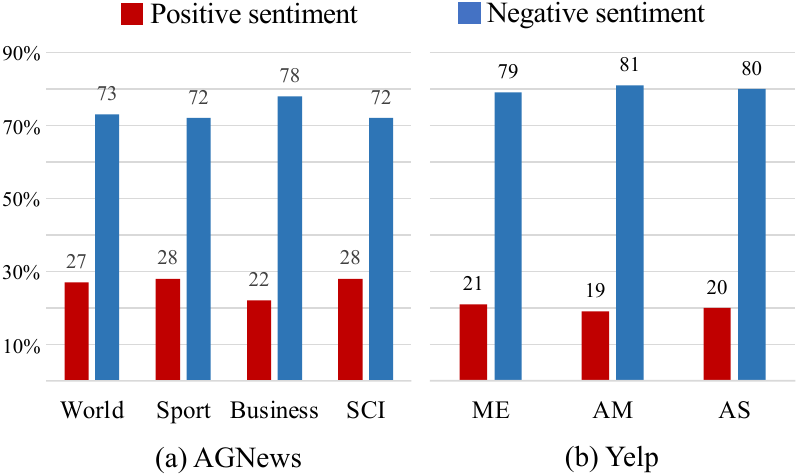}
\caption{The relevance scores of positive and negative sentiment in (a) AGNews and (b) Yelp. (a) The classifiers used for statistics are from \cite{discrete2022Gu}. (b) The statistical data of Yelp are from \cite{tailor2023Yang}.}
\label{fig:kia}
\end{figure}

Due to the difficulty of directly obtaining training data that satisfy arbitrary attribute combinations, existing methods \citep{GEDI2021Krause, Mac2023Ding, prior2023Gu} reuse datasets with single-aspect annotations for MCTG, where each training sample only expresses a single attribute in one aspect.
This neglects the fact that a sentence often couples multiple attributes due to the complexity of natural language.
The co-occurrence of attributes within one sentence forms patterns corresponding to attribute correlations, serving as crucial dependencies for a generative model in inference.
Meanwhile, the training corpus is derived from real life, where preferences in real life make certain combinations of attributes more common, leading to an imbalance in attribute correlations.
As an example shown in Figure~\ref{fig:kia}, in the AGNews dataset, since news with the topics of ``world'' and ``business'' are prone to correlating with negative elements \citep{discrete2022Gu}, such as war or inflation, combinations of these topics and negative sentiment are more prevalent.
In the Yelp dataset consisting of restaurant reviews with sentiment and food types, negative reviews also dominate \citep{tailor2023Yang}.
The imbalance in attribute correlations can lead the model to associate specific attributes, forming a stereotype that impacts multi-aspect control.
An MCTG model can better fit attributes with higher co-occurrence frequencies, allowing it to learn the semantic information of these attributes more comprehensively.
However, the model may neglect the learning of attributes with low co-occurrence frequencies, which hurts the control of these attribute combinations.

To resolve the problem, we propose a \textbf{m}ulti-\textbf{a}spect controllable text \textbf{g}eneration method with d\textbf{i}sentangled \textbf{c}ounterfactual augmentation, called \modelname.
Specifically, we introduce attribute disentanglement with latent space optimization.
It can disentangle the control factors of different attributes in the texts and generate the latent vectors with counterfactual features in the attribute latent space.
During training, we employ counterfactual latent vectors to balance attribute correlations, thereby constructing a more semantically balanced attribute latent space.
During inference, we enhance attribute correlations by the counterfactual latent vectors to improve multi-aspect control.

We experiment on three-aspect control including \textit{sentiment}, \textit{topic}, and \textit{detoxification}. 
We evaluate the relevance scores of attributes and assess the text quality in the scenarios of imbalanced and balanced attribute correlations. 
The results indicate that \modelname can leverage attribute correlations and mitigate the imbalance issues, which leads to steady and superior performance in both imbalanced and balanced scenarios than state-of-the-art baselines on multi-aspect control. 
We further demonstrate the effectiveness of each module in \modelname through analytical experiments. 

Our main contributions are outlined as follows:
\begin{compactitem}
    \item To mitigate the issue of imbalanced attribute correlations, we propose a counterfactual feature augmentation model with attribute disentanglement.

    \item To improve multi-aspect control by leveraging attribute correlations, we introduce a text generation method based on target-guided attribute correlation augmentation.

    \item We experimentally validate the effectiveness of \modelname.
    It outperforms state-of-the-art baselines on the imbalanced and balanced settings of multi-aspect control.
\end{compactitem}

\begin{figure*}
    \centering
    \includegraphics[width=\textwidth]{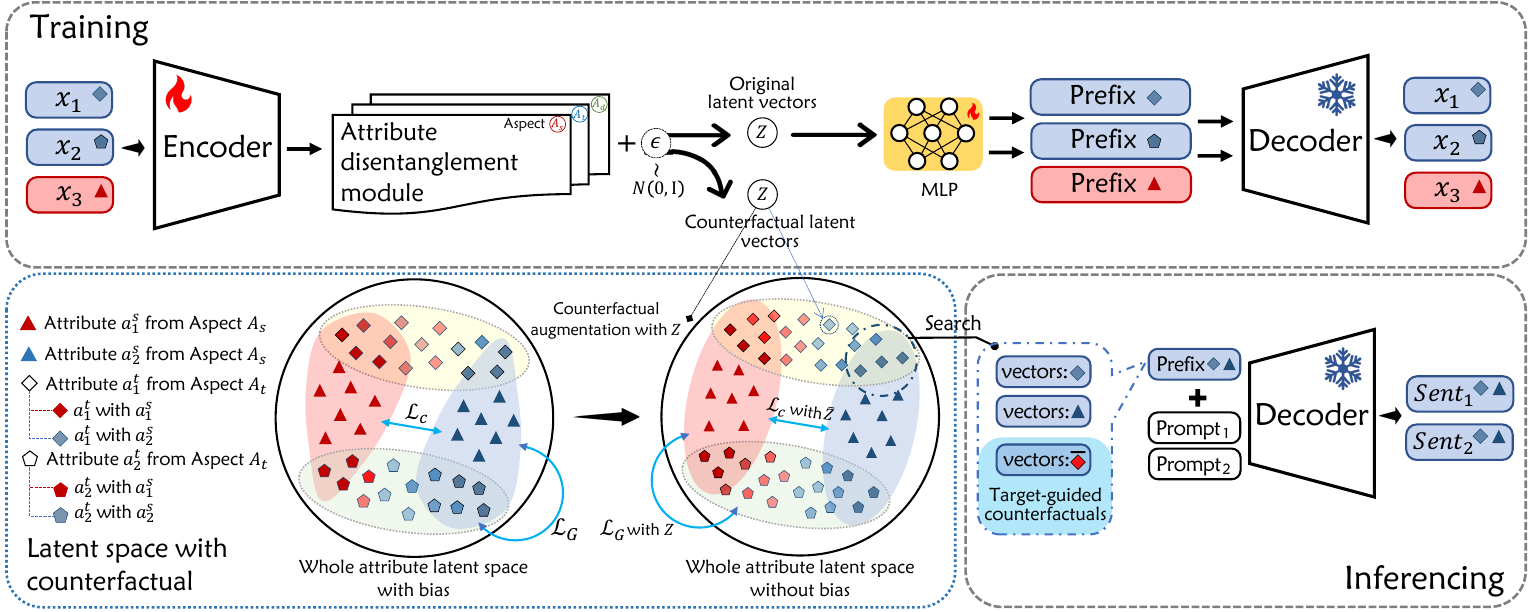}
    \caption{Framework of our method. Top part: We use the prefix tuning-based autoencoder structure as the framework and construct the attribute latent space. Bottom left: The vectors with counterfactual attribute features generated by the attribute disentanglement module are assisted in the construction of the attribute latent space. Bottom right:Inference stage with target-guided attribute correlation augmentation to improve multi-aspect control.}
    \label{fig:framework}
\end{figure*}

\section{Related Work}
\paragraph{Controllable text generation.}
LLMs introduce new perspectives for controllable text generation, such as post-processing \citep{GEDI2021Krause, Dexperts2021Liu} and prefix tuning \citep{2022qian, prefix2021li, 2021Yu}.
In contrast to single-aspect control, multi-aspect control has garnered increasing attention.
Current methods on MCTG can be broadly classified into three categories.
(i) \emph{Weighted decoding} is a kind of method that biases the output token distribution during decoding to achieve controllable generation \citep{PPLM2020Dath, fudge2021yang, Dexperts2021Liu, GEDI2021Krause, 2022Gu}.
(ii) The methods of \emph{optimization in the language space} model the generation of tokens satisfying multi-aspect requirements as a multi-objective optimization problem \citep{mix_match2022Mire, COLD2022qian, mucoco2021Kumar}.
(iii) The prefix tuning methods of \emph{optimization in the latent semantic space} have shown significant effectiveness in achieving multi-aspect control \citep{discrete2022Gu, OED2023Liu, tailor2023Yang, 2023Huang, prior2023Gu}.
However, these methods rely highly on the training data to construct the latent space and do not consider the influence of attribute correlations.

\paragraph{Counterfactual augmentation.}
Counterfactuals are designed to study the change in a response variable following an intervention.
Counterfactual augmentation is employed to enhance the robustness of models against the spurious correlations \citep{neurocounterfactuals2022Howard} including manual and automatic solutions.
The manual solution is a human-in-the-loop method to generate counterfactual texts by human annotators, which is costly and time-consuming \citep{2020Kaushik}.
For the automatic solution, some methods get counterfactual texts by finetuning LLMs \citep{polyjuice2021wu, Exploring2021yang, RGC2022Paranjape}.
Some methods propose controllable text generation approaches based on weighted decoding \citep{2021Madaan} or a structural causal model \citep{2021Hu}.
The above methods enhance the training set by generating counterfactual texts.
Following the above idea, we apply counterfactual augmentation to multi-aspect control and generate latent vectors with counterfactual features in the latent space to mitigate the impact of imbalanced attribute correlations.
We also leverage the attribute correlations promoted by counterfactuals to improve multi-aspect control.

\section{Formulation}
\paragraph{Task definition.}
Let $\mathbf{A}=\{A_{1}, \dots, A_{N}\}$ be $N$ aspects. 
Each aspect $A_{t} \in \mathbf{A}$ contains $|A_{t}|$ mutually exclusive attributes $\big\{a_{1}^{t}, \dots, a_{\left|A_{t}\right|}^{t}\big\}$.
The goal of MCTG is to generate sentences possessing multiple attributes from different aspects simultaneously.
For example, we may ask an MCTG model to generate texts with attribute ``sport'' from aspect \textit{topic}, attribute ``positive'' from aspect \textit{sentiment}, and attribute ``non-toxic'' from aspect \textit{detoxification}.

\paragraph{Training corpus.}
The training samples are indexed according to their associated aspects and attribute labels.
We denote the indices of all sentences with any attribute in aspect $A_{t}$ by $I^{t}$. 
The indices of the entire training data are $I=\bigcup_{t=1}^{N} I^{t}$.
Furthermore, $I_{a_{\mu}^t}^{t}$ is the indices of all training sentences about attribute $a_{\mu}^{t}$ in aspect $A_{t}$, and we have $I^{t}=\bigcup_{\mu=1}^{\left|A_{t}\right|} I_{a_{\mu}^{t}}^{t}$. 
Following \cite{focus2023Ma}, we introduce the concepts of explicit and implicit attributes to facilitate the explanation of our method.
For the data indexed by $I_{a_{\mu}^t}^{t}$, $a_{\mu}^{t}$ is the explicit attribute with annotated labels, and $A_{t}$ is the corresponding explicit aspect.
Other potential attributes in the data are the implicit attributes from implicit aspects, which are not explicitly provided but annotated by extra attribute classifiers in our work.
For notation, $I_{a_{1}^{t}, a_{1}^{k}}^{t}$ are the indices of data with the explicit attribute $a_{1}^{t}$ and the implicit attribute $a_{1}^{k}$.

\paragraph{Attribute correlation imbalance.}
Frequently co-occurred attributes tend to exhibit attribute correlations.
Once the co-occurrence frequency of an attribute pair significantly exceeds others, it exhibits imbalanced attribute correlations.
Given $I_{a_{\mu}^{t}}^{t}$,
suppose that we have $ I_{a_{\mu}^{t}}^{t} = I_{a_{\mu}^{t}, a_{\sigma}^{k}}^{t} + I_{a_{\mu}^{t}, a_{\bar{\sigma}}^{k}}^{t}$, where $a_{\sigma}^{k}$ and $a_{\bar{\sigma}}^{k}$ are two mutually exclusive implicit attributes from aspect $A_{k}$.
The attribute correlation becomes imbalanced when $\left| I_{a_{\mu}^{t} , a_{\sigma}^{k}}^{t} \right| \gg \left| I_{a_{\mu}^{t} , a_{\bar{\sigma}}^{k}}^{t} \right|$.

\paragraph{Counterfactual samples.}
Generally, a counterfactual sample is defined as a synthetically generated sentence that is treated differently by a condition model \citep{2021Madaan}.
Given a training sample $x$ with explicit attribute $a_{\mu}^{t}$ from aspect $A_{t}$ and an implicit attribute $a_{\sigma}^{k}$ from aspect ${A}_k$, we define the synthetically generated sample $\bar{x}$ with the same explicit attribute $a_{\mu}^t$ and reversed implicit attribute $a_{\bar{\sigma}}^k$ as a counterfactual sample to $x$.

\section{Methodology}
The structure of our \modelname is illustrated in Figure~\ref{fig:framework}.
Our \modelname employs an encoder-decoder framework based on prefix tuning to avoid the training costs in fine-tuning LLM.
We implement multi-aspect control revolving around the attribute latent space.
The encoder projects sampled sentences into the attribute latent space.
Several constraints are used to accurately model the attribute semantics.
We conduct resampling and incorporate latent vectors with counterfactual features to assist the construction of the attribute latent space and avoid the impact caused by the imbalanced attribute correlations.
The vectors with counterfactual features are generated by our designed attribute disentanglement module.
We also adopt a target-guided attribute correlation enhanced generation strategy to further improve multi-aspect control.

\subsection{Attribute Space Building Against Biases}
To facilitate multi-aspect control, we aim to construct an attribute latent space that models the semantics and relationships of attributes.
First, we need to establish a mapping between the attribute latent space and the attributes in sentences.
Various methods can be employed, such as VAE \citep{OED2023Liu, Mac2023Ding}.
Following \citep{discrete2022Gu}, we adopt a basic and simple method to map the attributes of sentences to discrete samples in the latent space.
Specifically, we leverage an encoder to extract the semantic features $\mathcal{H}_{i}$ in a sentence $x_{i}$: $\mathcal{H}_{i}=\operatorname{Encoder}\left(x_{i}\right)$. 
Then, we get latent vector ${Z}_{i}$ in the attribute latent space based on $\mathcal{H}_{i}$ through attribute disentanglement (described in Section~\ref{sec_disentanglement}).
We compute the prefix vector $\operatorname{Prefix}_{i}$ based on ${Z}_{i}$ in the attribute latent space as follows:
\begin{equation}
\operatorname{Prefix}_{i}=\operatorname{MLP}({Z}_{i}+\lambda \varepsilon), \label{eq:prefix}
\end{equation}
where $\lambda$ is a scaling factor and $\varepsilon$ is a perturbation vector from a multivariate Gaussian distribution for robustness.
The prefix is used to reconstruct the sentence $x_{i}$ and recover the corresponding attribute in the same way as an autoregressive loss:
\begin{equation}
\mathcal{L}_{Rec} = -\sum_{i \in I} \log p(x_{i}\mid\operatorname{Prefix}_{i}).
\label{eq:loss_rec}
\end{equation}

To accurately model attribute information, two kinds of constraints are utilized in modeling the attribute latent space \citep{discrete2022Gu, Mac2023Ding}:
(i) \emph{Classification loss} enables the differentiation of different attributes from the same aspect in the attribute latent space as follows:
\begin{align}
\mathcal{L}_{C} = - \sum_{t=1}^{|\mathbf{A}|}\sum_{\mu=1}^{|A_{t}|}\sum_{i \in I_{a_{\mu}^{t}}^{t}} \log p_{\pi_{t}}\big(a_{\mu}^{t} \mid {Z}_{i}\big),
\label{eq:LC}
\end{align}
where $p_{\pi_{t}}$ is a classifier to distinguish attribute $a_{\mu}^{t}$ among aspect $A_{t}$.
(ii) \emph{Aspect gap loss} aims to penalize the discrepancy of aspects caused by the domain gap among data sources and \addnew{facilitate the expression of multi-aspect semantics in the attribute latent space:}
\begin{align}
\resizebox{.89\columnwidth}{!}{$
\mathcal{L}_{G}=\sum\limits_{1 \leq t_{1}<t_{2}\leq|\mathbf{A}|}\Big\|\sum\limits_{i \in I^{t_{1}}} \frac{{Z}_{i}}{\left|I^{t_{1}}\right|} - \sum\limits_{j \in I^{t_{2}}} \frac{{Z}_{j}}{\left|I^{t_{2}}\right|}\Big\|_{2}^{2},
$}
\label{eq:LG}
\end{align}
\addnew{where $\| \cdot \|_{2}^{2}$ calculates the Euclidean distance based on the L2-norm.}

\begin{figure*}
    \centering
    \includegraphics[width=\textwidth]{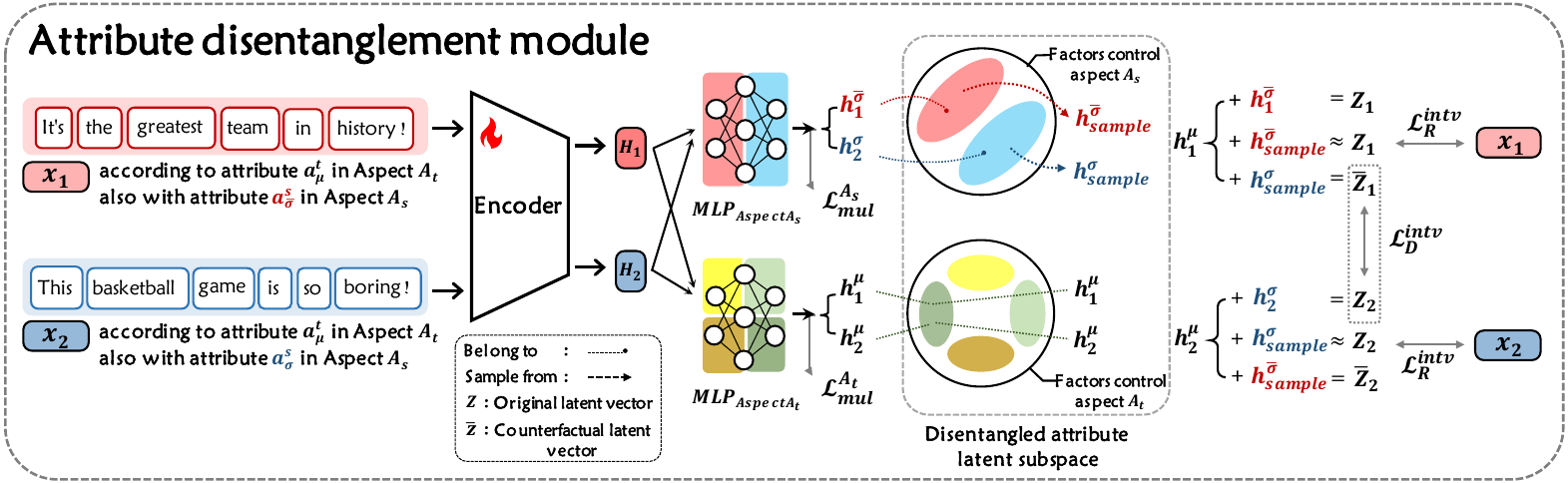}
    \caption{The attribute disentanglement module. $A_{t}$ and $A_{s}$ denote the explicit and implicit aspects, respectively.}
    \label{fig:Attribute_Disentanglement_Module}
\end{figure*}

When faced with imbalanced attribute correlations, samples of more frequently co-occurred attribute combinations are more likely to be selected during training.
The model is more prone to learning the semantics associated with these attribute combinations.
\addnew{Thus, we first adopt a resampling strategy to increase the probability of sampling the sentences with low-frequent attribute combinations. Specifically, when training with the data corresponding to each topic, for each sampled training example $x_{i}$, we simultaneously resample another example with the opposite sentiment to $x_{i}$. Furthermore, we also use the attribute disentanglement to generate latent vectors with counterfactual features to construct a more balanced attribute latent space.}
For aspect $A_{t}$ with imbalanced attribute correlations, the classification loss with counterfactual augmentation becomes
\begin{align}
\resizebox{\columnwidth}{!}{$
\mathcal{L}_{C}^{A_{t}} = - \sum\limits_{\mu=1}^{|A_{t}|}\sum\limits_{i\in I_{a_{\mu}^{t}}^{t}} \log\Big(p_{\pi_{t}} \big(a_{\mu}^{t} \mid {Z}_{i}\big)p_{\pi_{t}} \big(a_{\mu}^{t} \mid {\bar{Z}}_{i}\big)\Big),
$}
\label{eq:LC_Z}
\end{align}
and the aspect gap loss with counterfactual augmentation becomes
\begin{align}
\resizebox{.89\columnwidth}{!}{$
    \mathcal{L}_{G}^{A_{t}} = \sum\limits_{\substack{1\le t_{1} \le\left | A \right |\\ t_{1} \ne t }} \Big \| \sum\limits_{i\in I^{t} } \frac{Z_{i} + \bar{Z}_{i}}{2\times |I^{t}| } - \sum\limits_{j\in I^{t_{1}}} \frac{Z_{j}}{|I^{t_{1}}|} \Big \|_{2}^{2},
$}
\label{eq:LG_Z}
\end{align}
where ${\bar{Z}}_{i}$ is the latent vector with counterfactual features generated by the attribute disentanglement (described in Section~\ref{sec_disentanglement}).

\subsection{Attribute Disentanglement}
\label{sec_disentanglement}
In this section, we design an attribute disentanglement module to decouple the explicit and implicit attributes in sentences. 
Based on this, we can generate the latent vector $Z_{i}$ of the original sample $x_{i}$ and $\bar{Z}_{i}$ of the counterfactual sample $\bar{x}_{i}$ in the attribute latent space.
By transferring the shared implicit attribute features across the data with different explicit attributes, we can supplement the insufficient implicit attribute information in the data corresponding to each explicit attribute, caused by the low attribute co-occurrence frequency.

Figure~\ref{fig:Attribute_Disentanglement_Module} provides an overall description. 
Specifically, given $x_{i}$ with explicit attribute $a_{\mu}^{t}$ and implicit attribute $a_{\sigma}^{s}$, we refer to the latent vectors decoupled into the subspaces of explicit and implicit attributes as the respective explicit and implicit attribute control factors: $h_{i}^{\mu}$ and $h_{i}^{\sigma}$. 
Assuming $h_{i}^{\mu}=\operatorname{MLP}\big(\mathcal{H}_{i}\big)$, $h_{i}^{\sigma}$ are calculated similar to $h_{i}^{\mu}$ but with a different MLP, and $Z_{i}=h_{i}^{\mu} + h_{i}^{\sigma}$.

To implement the disentanglement of implicit and explicit attributes, we introduce three kinds of losses.
The first loss is \emph{Multi-task loss} which ensures that the control factors with attribute features extracted by explicit or implicit extractors can be discriminative for their respective attributes \citep{adv_dis2019John}.
Given the sentence $x_i$ needing disentanglement, the multi-task loss for each explicit or implicit aspect $A_{*}$ involved in the disentanglement is calculated like
\begin{align}
\mathcal{L}_{mul}^{A_{*}} =-\sum_{\beta=1}^{\left|A_{*}\right|} \sum_{i \in I_{a_{\beta}^{*}}^{*}} \log p_{\pi_{*}}\left(a_{\beta}^{*} \mid h_{i}^{\beta}\right),
\label{eq:ctrl_factors}
\end{align}
where $p_{\pi_{*}}$ is the classifier to distinguish attribute $a_{\beta}^{*}$ among aspect $A_{*}$, $h_{i}^{\beta}$ is the control factor of attribute $a_{\beta}^{*}$.

However, this cannot constrain the mutual influence between the control factors of explicit and implicit attributes.
Thus, we introduce two \emph{intervention losses}.
$\mathcal{L}_{R}^{intv}$ aims to eliminate the interference of the implicit attribute control factor on the explicit attribute and transfer the shared features of implicit attributes across different explicit attribute data.
Specifically, given a sentence $x_{i}$ with explicit attribute $a_{\mu}^{t}$ and implicit attribute $a_{\sigma}^{s}$, the respective control factors are $h_{i}^{\mu}$ and $h_{i}^{\sigma}$.
We first sample a sentence $x_{sample}$ with the same implicit attribute but a different explicit attribute compared to $x_{i}$, and denote its implicit attribute control factor by $h_{sample}^{\sigma}$.
We combine the explicit control factor $h_{i}^{\mu}$ with $h_{sample}^{\sigma}$ to get the prefix:
\begin{align}
\operatorname{ Prefix }_{i}^{intv} \! = \operatorname{MLP}\left(h_{i}^{\mu}+h_{\text {sample }}^{\sigma}\right).
\label{eq:pref_intv}
\end{align}

We ask the prefix to reconstruct a new sentence similar to $x_{i}$ (with the same explicit and implicit attributes as $x_{i}$).
The loss function is
\begin{align}
\resizebox{.89\columnwidth}{!}{$
\mathcal{L}_{R}^{intv} = - \sum\limits_{a_{\mu}^{t}\in A_{t}} \sum\limits_{i\in I_{a_{\mu}^{t}}^{t}}\log p_{L M}\left(x_{i} \mid \operatorname{Prefix}_{i}^{intv} \right).
$}
\label{eq:loss_intv}
\end{align}

This means that the implicit attribute control factors disentangled from any sentences with different explicit attributes do not affect the control of explicit attributes.
Meanwhile, the shared implicit attribute features in all sentences can be utilized for the reconstruction of each sentence without intervening in the explicit attributes.

However, relying solely on the above two losses is still insufficient.
\addnew{Without constraints, the explicit attribute control factor may still interfere with the implicit attributes.
Thus, we design the second loss, $\mathcal{L}_{D}^{intv}$, which aims to constrain the influence of explicit (e.g., topic) control factors on implicit attributes (e.g., sentiment).}
Specifically, given a sentence $x_{i}$ with explicit attribute $a_{\mu}^{t}$ and implicit attribute $a_{\sigma}^{s}$, the respective control factors are $h_{i}^{\mu}$ and $h_{i}^{\sigma}$.
We sample a sentence $x_{sample}$ with a different implicit attribute $a_{\bar{\sigma}}^{s}$ from $x_{i}$ and denote its implicit attribute control factor by $h_{sample}^{\bar{\sigma}}$. 
Then, we combine $h_{i}^{\mu}$ with $h_{sample}^{\bar{\sigma}}$ and instruct the model to reconstruct a sentence $\bar{x}_{i}$, which has the explicit attribute $a_{\mu}^{t}$ but changes the implicit attribute to $a_{\bar{\sigma}}^{s}$. 
\addnew{This means that we need to know what sentence $\bar{x}_{i}$ looks like after the change in sentiment. 
Since we do not have textual annotations for $\bar{x}_{i}$ and observe a tendency for text with similar attributes to exhibit cohesion in the attribute space, we attempt to enforce the constraints in the attribute space.}
Thus, we denote the vector of $\bar{x}_{i}$ in the attribute latent space by $\bar{Z}_{i}$, and bring $\bar{Z}_{i}$ closer to the set of vectors that share the same explicit and implicit attributes.
The loss function is
\begin{align}
\resizebox{.87\columnwidth}{!}{$
\begin{aligned}
\mathcal{L}_{D}^{intv} &= \sum_{a_{\mu}^{t} \in A_{t}} \sum_{i \in I_{a_{\mu}^{t}}^{t}} \max \Big(d\big(\bar{Z}_{i}, \hat{Z}_{i}\big) - \gamma, 0\Big), \\
\bar{Z}_{i} &= h_{i}^{\mu}+h_{\text {sample }}^{{\bar{\sigma}}}, \\
\hat{Z}_{i} &= \frac{1}{\left|I_{a_{\mu}^{t}, a_{\bar{\sigma}}^{s}}^{t}\right|} \sum_{j \in I^{t}_{a_{\mu}^{t}, a_{\bar{\sigma}}^{s}}} h_{j}^{\mu}+h_{j}^{\bar{\sigma}}.
\end{aligned}
$}
\label{eq:loss_intv2}
\end{align}

If the explicit attribute control factor affects the implicit attribute, it would impact the modification of implicit attributes after replacing the implicit attribute control factor. 
We constrain the potential impact through the loss $\mathcal{L}_{D}^{intv}$, while ensuring the function of the implicit attribute control factor.

\subsection{Multi-Aspect Generation}
Suppose the target combination of attributes $A_{\text{target}}$ is $\left\{a_{\varphi_{1}}^{1}, \dots, a_{\varphi_{N}}^{N}\right\}$ from $N$ different aspects, where $a_{\varphi_{t}}^{t}$ is the $\varphi_{t}$-th attribute from aspect $A_{t}$.
We implement multi-aspect control revolving around the attribute latent space.
Since the attribute correlations benefit the control of the corresponding attribute combinations, we adopt a multi-aspect generation strategy with target-guided attribute correlation augmentation.
Specifically, we first utilize the control factors of explicit and implicit attributes from attribute disentanglement to generate latent vectors in the attribute space aligning with the target attribute combinations.
Then, we use an iterative intersection retrieval algorithm in the attribute latent space to get the latent vector that simultaneously satisfies the target attributes following \citep{discrete2022Gu}.
For each target attribute $a_{\mu}^{t}$, we identify the top K vectors among the set of vectors corresponding to $a_{\mu}^{t}$, which are closest to the vectors corresponding to other attributes in $A_{\text{target}}$.
The mean value of the top K vectors is used to represent the corresponding attribute.
We calculate the weighted sum of each representative vector to obtain the target vector:
\begin{align}
\resizebox{.87\columnwidth}{!}{$
\tilde{Z}=\sum\limits_{a_{\mu}^{t} \in A_{\text{target}}} w_{a_{\mu}^{t}} \times \operatorname{mean}\Big(Z_{i}, i\in\operatorname{N}_{\text{topK}} \big(I_{a_{\mu}^{t}}^{t}\big)\Big),
$}
\end{align}
where $\operatorname{mean}(\cdot)$ is the mean operation within the set, $\operatorname{N}_{\text{topK}}\big(I_{a_{\mu}^{t}}^{t}\big)$
are the indices of the top K vectors for the current attribute $a_{\mu}^{t}$ and are closest to the vectors of other attributes under control.
We use $\tilde{Z}$ to get the prefix by Eq.~\ref{eq:prefix} and generate the target sentence based on the prefix and prompt $\tilde{x}$:
\begin{align}
Y=\mathop{\arg\max}_{y} p_{\mathrm{LM}}\big(y \mid {\text{Prefix}};\tilde{x}\big).
\end{align}

\begin{table*}[!t]
\centering
\small{
\begin{tabular}{c|l|c|ccc|c|c}
\toprule
\multicolumn{2}{c|}{Methods} & Avg. $\uparrow$ ($\%$) & Sentiment $\uparrow$ ($\%$) & Topic $\uparrow$ ($\%$) & Detoxification $\uparrow$ ($\%$) & PPL $\downarrow$ & Distinct $\uparrow$\\
\midrule
\multirow{8}{*}{\rotatebox[origin=c]{90}{Imbalanced}} \multirow{8}{*}{\rotatebox[origin=c]{90}{attribute correlations}} 
& \addnew{PPLM}     & \addnew{70.7 $\pm$ 24.9} & \addnew{63.6 $\pm$ 28.7} & \addnew{61.8 $\pm$ 25.9} & \addnew{86.9 $\pm$ 9.5} & \addnew{69.8} & \addnew{60.2} \\ 
& GeDi              & 82.3 $\pm$ 18.6 & 73.5 $\pm$ 23.1 & 77.8 $\pm$ 16.9 & 95.5 $\pm$ 2.6 & 92.2 & 78.2 \\ 
& Mix\&Match        & 77.7 $\pm$ 22.7 & 72.5 $\pm$ 27.8 & 68.7 $\pm$ 23.6 & 91.8 $\pm$ 2.5 & 73.9 & 59.3 \\ 
& Tailor            & 76.9 $\pm$ 24.9 & 67.5 $\pm$ 31.3 & 66.7 $\pm$ 19.8 & \textbf{96.4} $\pm$ 1.9 & 26.8 & 69.8 \\ 
& LatentOPs         & 82.8 $\pm$ 16.2 & 78.1 $\pm$ 20.3 & 78.2 $\pm$ 15.4 & 92.1 $\pm$ 8.2 & 11.7 & 39.7 \\ 
& Discrete          & 83.8 $\pm$ 20.7 & 91.2 $\pm$ 15.6 & 65.5 $\pm$ 23.9 & 94.8 $\pm$ 3.6 & 43.1 & 42.1 \\ 
& MacLaSa           & 84.7 $\pm$ 13.9 & 82.4 $\pm$ 13.7 & 77.9 $\pm$ 16.8 & 93.9 $\pm$ 3.3 & 29.3 & 59.7 \\ 
& PriorControl      & 86.2 $\pm$ 13.6 & 88.1 $\pm$ 10.3 & 78.4 $\pm$ 19.2 & 92.1 $\pm$ 4.2 & 34.1 & 51.8 \\ 
& \modelname (ours) & \textbf{92.6} $\pm$ \ \,9.1  & \textbf{94.5} $\pm$ \ \,6.9 & \textbf{88.5} $\pm$ 13.4 & 94.7 $\pm$ 3.9 & 43.4 & 53.3 \\
\midrule
\multirow{8}{*}{\rotatebox[origin=c]{90}{Balanced}} \multirow{8}{*}{\rotatebox[origin=c]{90}{attribute correlations}}\
& \addnew{PPLM}     & \addnew{71.0 $\pm$ 21.4} & \addnew{64.7 $\pm$ 24.8} & \addnew{63.5 $\pm$ 22.7} & \addnew{84.9 $\pm$ 6.5} & \addnew{62.6} & \addnew{62.0} \\ 
& GeDi            & 81.4 $\pm$ 14.7 & 76.1 $\pm$ 17.2 & 73.8 $\pm$ 11.3 & 94.2 $\pm$ 1.9 & 116.6\ \, & 75.1 \\ 
& Mix\&Match      & 79.7 $\pm$ 21.8 & 73.5 $\pm$ 25.9 & 69.9 $\pm$ 21.1 & 95.8 $\pm$ 1.9 & 63.0 & 61.8 \\ 
& Tailor          & 78.1 $\pm$ 22.6 & 64.6 $\pm$ 28.5 & 73.7 $\pm$ 16.5 & \textbf{95.9} $\pm$ 2.5 & 28.7 & 69.8 \\ 
& LatentOPs       & 85.5 $\pm$ 14.4 & 76.3 $\pm$ 16.4 & 85.1 $\pm$ 14.1 & 94.9 $\pm$ 4.2 & 16.8 & 41.3 \\ 
& Discrete        & 87.4 $\pm$ 10.9 & 86.7 $\pm$ 10.5 & 84.8 $\pm$ 14.2 & 90.7 $\pm$ 7.4 & 28.4 & 49.5 \\ 
& MacLaSa         & 88.2 $\pm$ 10.7 & 85.0 $\pm$ 14.7 & 85.1 $\pm$ \ \,9.5 & 94.5 $\pm$ 2.6 & 19.2 & 56.5 \\ 
& PriorControl    & 92.2 $\pm$ \ \,8.6  & 92.5 $\pm$ \ \,8.5 & 89.3 $\pm$ 11.0 & 94.9 $\pm$ 3.4 & 29.6 & 51.6 \\ 
& \modelname (ours)            & \textbf{92.9} $\pm$ \ \,8.5  & \textbf{94.2} $\pm$ \ \,6.4 & \textbf{89.4} $\pm$ 12.2 & 95.1 $\pm$ 4.9 & 55.3 & 52.2 \\
\bottomrule
\end{tabular}
}
\caption{Automatic results of multi-aspect control with imbalance and balance attribute correlation. The best relevance scores are marked in \textbf{bold}. More results are shown in Tables~\ref{tab:detail_result_imb} and~\ref{tab:detail_result_balance} in the appendix.}
\label{tab:tb1}
\end{table*}

\section{Experiments and Results}
\subsection{Experiment Setup}
\paragraph{Datasets.}
We experiment with three-aspect control: \textit{sentiment}, \textit{topic}, and \textit{detoxification}.
Following previous works \citep{GEDI2021Krause, discrete2022Gu, prior2023Gu}, we pick IMDb for \textit{sentiment}, AGNews for \textit{topic}, and the Jigsaw Toxic Comment Classification Challenge dataset for \textit{detoxification}.
We simulate the imbalanced setting on the AGNews dataset.
For each topic in AGNews, the proportions of sentences with negative and positive sentiment are set to 7:3.
For the balanced setting, we reuse the dataset in \cite{discrete2022Gu}.
\addnew{More details are provided in Appendix~\ref{app:Datasets}.}

\paragraph{Baselines.}
We compare \modelname to 8 representative and strong baselines. 
(i) \textit{Weighted decoding}: \addnew{PPLM \cite{PPLM2020Dath} and GeDi \citep{GEDI2021Krause} bias the decoding process in generation.} 
(ii) \textit{Optimization in the language space}: Mix\&Match \citep{mix_match2022Mire} discretely optimizes sentences in the language space by token-level masking. 
(iii) \textit{Optimization in the latent space}: Tailor \citep{tailor2023Yang} is based on soft prompt-tuning. 
Discrete \citep{discrete2022Gu} uses discrete samples to construct the attribute latent space. 
LatentOPs \citep{OED2023Liu} adopts an efficient sampler based on ordinary differential equations (ODEs). 
MacLaSa \citep{Mac2023Ding} combines VAE and ODEs for the generation.
PriorControl \citep{prior2023Gu} utilizes the normalizing flow to constrain the complex latent space.
See Appendix~\ref{app:hyper} for implementation details.

\paragraph{Evaluation metrics.}
We compute the attribute relevance with the DeBERTa classifiers for \textit{sentiment} and \textit{topic} aspects.
We measure the \textit{non-toxicity} aspect with the Google Perspective API.
We consider two auxiliary metrics for text quality, i.e., perplexity (abbr. PPL) and distinctness.
\addnew{More details are provided in Appendix~\ref{app:Evaluation}.}
We also conduct human evaluations on the generated sentences.
Details and results are provided in Appendix~\ref{app:human_eval}.

\subsection{Main Results}
We conduct experiments on both imbalanced and balanced attribute correlation settings. 
Table~\ref{tab:tb1} lists the results. 
We report the average scores with standard deviations of 8 combinations for each aspect, as well as the average scores for all three aspects.

Overall, most methods perform worse in the imbalanced setting compared to the balanced one. 
This is due to the dominant negative impact of stereotypes formed by imbalanced attribute correlations.
The stereotypes hinder the classifiers used to optimize in the language space for Mix\&Match. 
The lack of data with positive sentiment combinations also affects the learning of semantics related to positive sentiment for each topic in the attribute latent space, such as Discrete and PriorControl.

\begin{table}[!t]
\centering
\resizebox{\columnwidth}{!}{
\begin{tabular}{l|c|ccc}
\toprule
Variants (strategies during training) & Avg. & Sent. & Topic & Detox. \\
\midrule
\modelname (intact)                      & 92.6 & 94.5 & 88.5 & 94.7 \\
~ w/o counterfactual (Eqs.~\ref{eq:LC_Z} and~\ref{eq:LG_Z})       & 90.5 & 91.5 & 86.1 & 94.1 \\
~ w/o resample strategies                & 88.5 & 88.9 & 83.1 & 93.7 \\
~ \addnew{w/o $\mathcal{L}_{C}$ (Eq.~\ref{eq:LC})}  & \addnew{85.9} & \addnew{89.0} & \addnew{76.5} & \addnew{92.1} \\
~ \addnew{w/o $\mathcal{L}_{G}$ (Eq.~\ref{eq:LG})}  & \addnew{88.6} & \addnew{90.6} & \addnew{81.0} & \addnew{94.3} \\
~ \addnew{w/o $\mathcal{L}_{mul}^{A_{*}}$ (Eq.~\ref{eq:ctrl_factors})} & \addnew{91.3} & \addnew{91.7} & \addnew{86.5} & \addnew{95.6} \\ 
~ \addnew{w/o $\mathcal{L}_{R}^{intv}$ (Eq.~\ref{eq:loss_intv})}    & \addnew{84.6} & \addnew{79.1} & \addnew{80.7} & \addnew{94.1} \\
~ \addnew{w/o $\mathcal{L}_{D}^{intv}$ (Eq.~\ref{eq:loss_intv2})}   & \addnew{83.0} & \addnew{81.1} & \addnew{75.5} & \addnew{92.5} \\
\midrule
Variants (strategies during inference) & Avg. & Sent. & Topic & Detox. \\
\midrule
\modelname (intact)                  & 92.6 & 94.5 & 88.5 & 94.7 \\
~ w/o all             & 86.9 & 85.8 & 79.1 & 95.8 \\
~ w/ balance          & 87.8 & 86.9 & 82.5 & 93.9 \\
\bottomrule
\end{tabular}
}
\caption{Analysis of different strategies.}
\label{tab:analysis_in_effects}
\end{table}

Due to the strategies used in training, \modelname is less affected by the imbalanced attribute correlations.
In the imbalanced setting, \modelname performs best on average attribute-related metrics, showing a 7.4\% improvement beyond the second-best method PriorControl.
The advances come from the improvement of \textit{topic} (12.8\%) and \textit{sentiment} (7.2\%) aspects. 
The target-guided counterfactual augmentation makes \modelname achieve better performance. 
Thus, in the balanced setting, \modelname can also achieve comparable performance with PriorControl  (1.8\% improvement in the \textit{sentiment} aspect).

In addition, GeDi performs well on attribute relevance and diversity, while badly on perplexity. 
\modelname exhibits a slightly higher PPL but still in a reasonable range compared with GeDi and Mix\&Match. \addnew{The disentanglement module tends to bring the latent vectors with similar features closer within one aspect, which can increase the distances between different aspects, potentially affecting the distribution of the normal attribute space.} 


\begin{table}[!t]
\centering
\resizebox{.9\columnwidth}{!}{
\begin{tabular}{l|cccc|c}
\toprule
\multirow{2}{*}{Topics} & \multicolumn{4}{c|}{Change factor of sent. from} & \multirow{2}{*}{Avg.} \\
\cmidrule(lr){2-5} & World & Sport & Business & Tech & \\
\midrule
World     & 90.7 & 89.2 & 74.9 & 86.1 & 83.4 \\
Sport     & 97.9 & 98.7 & 97.9 & 98.2 & 98.0 \\
Business  & 84.3 & 84.6 & 87.8 & 86.3 & 85.1 \\
Tech      & 98.6 & 99.3 & 99.4 & 99.5 & 99.1 \\
\bottomrule
\end{tabular}
}
\caption{Relevance scores of topic after changing the control factor of sentiment from different topics.}
\label{tab:disentangle1}
\end{table}

\subsection{Further Analysis}
\paragraph{Effects of different strategies.}
We validate the effects of strategies during training.
Table~\ref{tab:analysis_in_effects} lists the results.
After removing counterfactual augmentation, all relevance scores of our \modelname decrease.
We further remove the resampling strategy and also observe a decrease in performance.
We find that the counterfactual augmentation and resampling both make sense for the construction of attribute latent space.
\addnew{The loss functions of $\mathcal{L}_{C}$ (Eq.~\ref{eq:LC}) and $\mathcal{L}_{G}$ (Eq.~\ref{eq:LG}) are commonly used in previous works \citep{discrete2022Gu, Mac2023Ding} and both contribute to the performance. Our proposed loss functions $\mathcal{L}_{mul}^{A_{*}}$, $\mathcal{L}_{R}^{intv}$, and $\mathcal{L}_{D}^{intv}$ affect the performance by influencing disentanglement. After removing each of them, the decline in the effect of disentanglement also results in a decrease in performance.} 

In inference, we use a target-guided attribute correlation augmentation strategy to improve multi-aspect control, which generates new latent vectors in the attribute latent space aligning with the attribute combinations that we control.
We compare three variant strategies in Table~\ref{tab:analysis_in_effects}.
``w/o all'' is the variant without any augmentation strategy, which performs worst.
``w/ balance'' maintains an equal number of vectors for different attribute combinations in the attribute latent space.
We can find that the target-guided strategy used in \modelname is more effective than the balanced variant.

\paragraph{Analysis of attribute disentanglement.}
We perform experiments about attribute disentanglement.
First, we measure the impact of the disentangled implicit attribute (\textit{sentiment}) control factors on the explicit attribute (\textit{topic}).
Table~\ref{tab:disentangle1} lists the results.
For each topic in each row of the table, we generate texts by replacing the disentangled sentiment control factor with those corresponding to the other three topics.
The diagonal of the table represents generated texts using the original topic and sentiment control factors without replacement.
We measure the topic relevance scores of the generated texts.
It can be observed that, after replacing the sentiment factors with those from other topics, the topic relevance scores only show a slight decrease.

\begin{figure}
\centering
\includegraphics[width=\columnwidth]{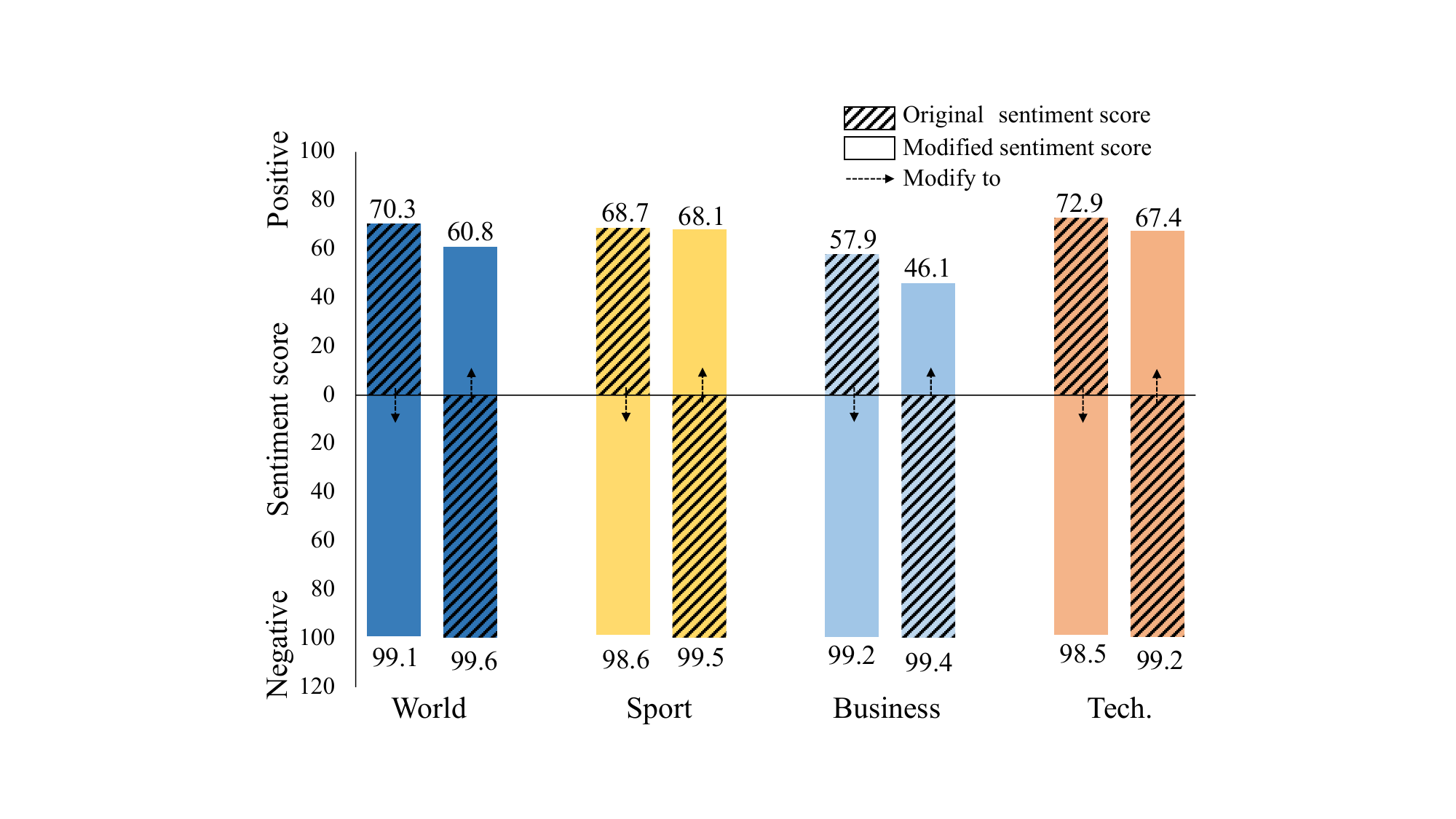}
\caption{Relevance scores of \textit{sentiment} after changing the control factor of sentiment to the opposite.}
\label{fig:attr_dis2}
\end{figure}

Next, we validate whether the disentangled explicit control factor (\textit{topic}) interferes with the implicit control factor in controlling implicit attributes (\textit{sentiment}).
The results are shown in Figure~\ref{fig:attr_dis2}. 
For each topic, the hatched bar chart represents the sentiment scores of the texts generated by the sentences with positive/negative sentiments in that topic (denoted by the original sentiment score).
The solid bar chart, opposite to the hatched bar chart, represents the sentiment scores of sentences generated after replacing the sentiment control factors with the opposite sentiment of the original sentences (denoted by the modified sentiment score).
Since we train the disentanglement module based on existing sentiment labels in each topic, the original sentiment scores serve as an upper bound for the modified sentiment scores of the given sentiment.
We can observe that for all topics, the original sentiment scores for negative sentiment are higher compared to positive sentiment. 
This is due to the biases formed by the decoder during pre-training.
In addition, it is easier to convert positive sentiment into negative sentiment as more sentences with negative sentiments in the training set providing more supervision.
The conversion of negative sentiment to positive sentiment is less effective, which we ascribe to the limited availability of supervision signals for positive sentiment. 

\addnew{Since we utilize $\mathcal{L}_{mul}^{A_{*}}$, $\mathcal{L}_{R}^{intv}$, and $\mathcal{L}_{D}^{intv}$ for attribute disentanglement, we also introduce ablation studies to investigate the impact of these loss functions. Details are provided in Appendix~\ref{secs:ablation_disentanglement}.}

\begin{figure}
\centering
\includegraphics[width=.99\columnwidth]{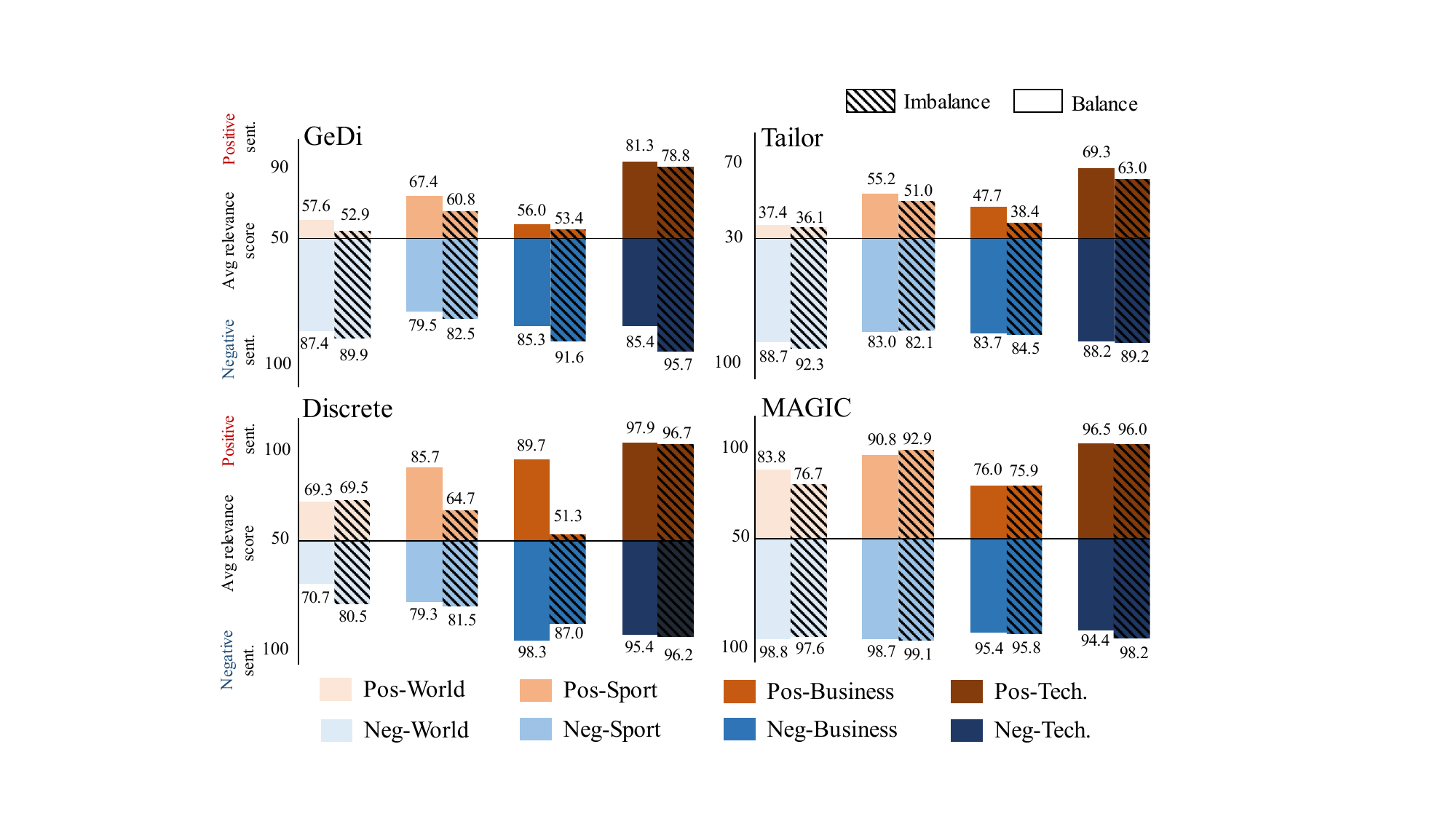}
\caption{Effects of attribute correlation imbalance in performance with different attribute combinations.}
\label{fig:ana_2}
\end{figure}

\paragraph{Effects of attribute correlation imbalance.}
We analyze the effects of attribute correlation imbalance in performance with different attribute combinations. 
Figure~\ref{fig:ana_2}  illustrates the performance of \textit{topic} and \textit{sentiment} combinations under balanced and imbalanced attribute correlations.
For each method, the upper/lower side of the x-axis corresponds to the average attribute scores for the 4 combinations of positive/negative sentiment and 4 different topics.
The hatched bar chart represents the scenario of imbalance attribute correlation. 
It can be observed that the performance of the majority of positive sentiment attribute combinations shows varying degrees of decline under the imbalanced setting.  
This is due to the limited samples of positive sentiment attribute combinations in the training data.
Certain methods exhibit an improvement in the performance of negative sentiment attribute combinations in the imbalanced setting, such as GeDi and Discrete.
\modelname is least affected due to the utilization of balancing strategies during training and the target-guided counterfactual augmentation of generation during inference.

\section{Conclusion}
In this paper, we consider attribute correlation and propose a novel method, \modelname, for multi-aspect control with disentangled counterfactual augmentation.
We alleviate the issue of imbalanced attribute correlations during training and further improve multi-aspect control using counterfactual vectors in the attribute latent space by disentanglement. Experiments on the three-aspect control task support the effectiveness of \modelname. We also conduct detailed analytical experiments to study the effects of each strategy in \modelname. In the future, we will explore the impact of attribute correlations formed during pre-training.

\section*{Ethical Considerations}
The training data used in our work are sourced from the web and have not undergone extensive data cleansing. As a result, the method that we propose and the baselines to which we compare may produce some fake, toxic, or offensive content. It is important to clarify that the generated texts in our work do not represent our viewpoints. Additionally, detoxification is considered as a default attribute that the generated texts are expected to satisfy. We believe that exploring controllable generation techniques is beneficial for combating the generation of harmful texts.

\section*{Limitations}
\modelname has several limitations:
(i) To construct the attribute latent space, our method requires a substantial amount of training data, making it challenging to address the few-shot scenario. 
(ii) For attribute disentanglement, \modelname needs an extra pre-trained classifier for labeling implicit attributes. The performance of this classifier could potentially impact the effectiveness of disentanglement.
In the future, we will explore strategies to reduce reliance on classifiers for the disentanglement of control attributes.
(iii) For a fair comparison, our decoder has a moderate parameter count, such as GPT2-medium. 
In the future, exploring more complex controllable generation tasks with a larger LLM would also be interesting.

\section*{Acknowledgments}
We thank the anonymous reviewers for their valuable comments.
This work was supported by the National Natural Science Foundation of China (No. 62272219) and the Collaborative Innovation Center of Novel Software Technology \& Industrialization.

\bibliography{anthology,custom}

\appendix

\begin{table*}[!t]
\centering
\addnew{
\small{
\begin{tabular}{l|c|ccc|c|c}
\toprule
Methods & Avg. $\uparrow$ ($\%$) & Sentiment $\uparrow$ ($\%$) & Topic $\uparrow$ ($\%$) & Detoxification $\uparrow$ ($\%$) & PPL $\downarrow$ & Distinct $\uparrow$\\
\midrule
ChatGPT    & 92.5 $\pm$ 11.7  & \textbf{97.4} $\pm$ 3.1 & 83.5 $\pm$ 17.1 & \textbf{96.6} $\pm$ 2.6 & 18.7 & 53.9 \\ 
\modelname (ours) & \textbf{92.6} $\pm$ \ 9.1  & 94.5 $\pm$ 6.9 & \textbf{88.5} $\pm$ 13.4 & 94.7 $\pm$ 3.9 & 43.4 & 53.3 \\
\bottomrule
\end{tabular}
}}
\caption{\addnew{Automatic results on multi-aspect control compared with ChatGPT. More results are shown in Tables~\ref{tab:detail_chatgpt}}.
}
\label{tab:chatgpt}
\end{table*}

\section{Details of Hyperparameter Selection}
\label{app:hyper}
We describe the hyperparameter selection of our method in this section.
The implementation of the encoder and decoder follows the previous works \citep{discrete2022Gu, prior2023Gu, Mac2023Ding}.
The encoder is initialized with Bert-base-uncased and subsequently finetuned during training.
The decoder uses GPT2-medium\footnote{\url{https://huggingface.co/openai-community/gpt2-medium}} and fixed.
Each sentence, after being encoded and mean-pooled, is converted to a 768-dimensional latent representation. 
The latent representations are mapped to the prefix with a dimension of (20, 24, 2, 1,024), where 20 is the prefix sequence length, 24 is the number of hidden layers in GPT2-medium, 2 represents one key and one value, and 1,024 is the size of hidden states in GPT2-medium.
The dimension of the multi-layer perceptrons used in the attribute disentanglement is $768 \times 768$.

During the training stage, we use half-precision mode for efficiency on one NVIDIA A800 80GB GPU, where the batch size is 64.
The scaling factors of $\mathcal{L}_{Rec}$, $\mathcal{L}_{C}$, $\mathcal{L}_{G}$, $\mathcal{L}_{mul}^{A_{*}}$, $\mathcal{L}_{R}^{intv}$, and $\mathcal{L}_{D}^{intv}$ are 0.5, 0.2, 0.3, 0.2, 0.5, and 0.2, respectively. 
The margin $\gamma$ in $\mathcal{L}_{D}^{intv}$ is 0.4. 
The optimizer is AdamW with a learning rate of 1e-4, the number of training epochs is 300, and we use a checkpoint at a step of 57,000.

In the inference, we use the initial settings as \citep{discrete2022Gu} for the iterative intersection retrieval algorithm. 
We employ grid search with for loops to search for the optimal weight parameters for each combination of attributes, which aims to balance the performance among attributes from different aspects. 
The search range of weights for sentiment is \{1.5, 2.5, 3.5\}, while the range for the topic is 2.0 to 10.5 with an interval of 1.5. 
The weight of detoxification is set to 1.0. 
The generation process is the same as prefix tuning and the length of the generated text is set to 50.

Since our method needs an extra attribute classifier for the implicit attributes, we use the DeBERTa-based model to train a classifier for sentiment. 
We sample 100k, 10k, and 10k sentiment-specific sentences from the Yelp dataset for training, validation, and testing, respectively. 
The learning rate is set to 5e-5 and the batch size is set to 64. 
The F1 score of the classifier on the testing set is 97.09. 

Detailed settings of baselines are as follows: (i) For the weighted decoding method GeDi, we directly train the classifiers of three aspects on our datasets.
(ii) For the multi-objective optimization method Mix\&Match, we follow the experiment setting in the previous work. 
We retrain the classifiers and use sentences generated from PPLM \citep{PPLM2020Dath} to initialize the sampling process so that long sentences can be generated for a fair comparison.
(iii) For the methods optimizing in latent space, we reproduce LatentOps and MacLaSa by retraining the classifiers and the VAE module on our datasets. We retrain the soft prompts of Tailor on our datasets following their default hyperparameters. For Discrete and PriorControl, we utilize the same way as our method to select the optimal weight parameters, and other hyperparameters are kept default.
We uniformly use the pre-trained language model to GPT2-medium except for Mix\&Match using Bert-large.\footnote{\url{https://huggingface.co/google-bert/bert-large-uncased}}

\section{Details of Experiment Setup}
\addnew{
\subsection{Datasets}
\label{app:Datasets}
Following previous works \citep{GEDI2021Krause, discrete2022Gu, prior2023Gu}, we pick IMDb \citep{imdb2011Maas} for \textit{sentiment}, AGNews for \textit{topic} \citep{agnews2015Zhang}, and the Jigsaw Toxic Comment Classification Challenge dataset for \textit{detoxification}, with 10k samples for each aspect and equal samples for each attribute.
Using 35 prompts from \citep{PPLM2020Dath}, we assess 8 combinations across 2 sentiments, 4 topics, and 1 detoxification, generate 5 sentences per combination, and evaluate 1,400 sentences in total.
\subsection{Evaluation Metrics}
\label{app:Evaluation}
We compute the attribute relevance with the DeBERTa classifiers finetuned on the Yelp and AGNews datasets~\citep{agnews2015Zhang} for \textit{sentiment} and \textit{topic} aspects, respectively.
Both classifiers are from \cite{prior2023Gu}.
The \textit{non-toxicity} aspect is measured by the Google Perspective API.\footnote{\url{https://www.perspectiveapi.com}}
We also consider two auxiliary metrics for text quality, i.e., perplexity (abbr. PPL) and distinctness.
PPL is calculated by GPT2-large and distinctness is calculated by averaging the 1-gram, 2-grams, and 3-grams distinct scores \citep{tailor2023Yang}.}

\begin{table*}[!t]
\centering
\addnew{
\small{
\begin{tabular}{l|cc|cccc|c}
\toprule
\multirow{2}{*}{ChatGPT} & \multicolumn{2}{c|}{Sentiment ($\%$)} & \multicolumn{4}{c|}{Topic ($\%$)} & \multirow{2}{*}{Detox. ($\%$)} \\
\cmidrule(lr){2-3} \cmidrule(lr){4-7} 
& Neg. & Pos. & World & Sport & Business & Tech. & \\
\midrule

Comb. 1 & 91.4 & - & 67.2 & - & - & - & 95.6\\
Comb. 2 & 95.0 & - & - & 96.3 & - & - & 92.9\\
Comb. 3 & 96.1 & - & - & - & 86.0 & - & 94.2\\
Comb. 4 & 97.2 & - & - & - & - & 99.2 & 94.3\\
Comb. 5 & - & 99.9 & 54.6 & - & - & - & 98.9\\
Comb. 6 & - & 99.7 & - & 94.9 & - & - & 99.0\\
Comb. 7 & - & 99.9 & - & - & 70.5 & - & 99.4\\
Comb. 8 & - & 99.9 & - & - & - & 98.9 & 98.9\\
\cmidrule(lr){1-8}
Avg.  & 94.9 & 99.9 & 60.9 & 95.6 & 78.3 & 99.1 & 96.7\\
\bottomrule
\end{tabular}
}}
\caption{\addnew{Detailed results of ChatGPT on multi-aspect control.}}
\label{tab:detail_chatgpt}
\end{table*}

\section{Statistics in Figure~\ref{fig:kia}}
For Yelp, we directly utilize the statistical results from \citep{tailor2023Yang}, which filter out neutral texts. More details can be found in the relevant paper. 
For AGNews, unfiltered neutral texts exist in the original dataset. 
We manually select some neutral texts and adjust the temperature of the classifier to keep the sentiment scores of neutral texts around 0.5, thereby mitigating their impact on the statistical results.
The temperatures for the four topics (``world'', ``sport'', ``business'', ``technology'') are 6.5, 4.5, 5, and 4.5, respectively.

\begin{table*}[!t]
\centering
\addnew{
\small{
\begin{tabular}{l|cccc|cc}
\toprule
Original control factors & World & Sport & Business & Tech. & Positive & Negative \\
\midrule
\modelname (intact) & 90.7 & 98.7 & 87.8 & 99.5 & 67.5  & 99.4 \\
\toprule
Change control factors & World & Sport & Business & Tech & Positive & Negative \\
\midrule
\modelname (intact) & 83.4\ (7.3$\downarrow$) & 98.0\ (0.7$\downarrow$) & 85.1\ (2.7$\downarrow$) & 99.1\ (0.4$\downarrow$) & 60.6\ (6.9$\downarrow$) & 98.8\ (0.6$\downarrow$) \\
\rule{0pt}{10pt}
~ w/o $\mathcal{L}_{mul}^{A_{*}}$ (Eq.~\ref{eq:ctrl_factors}) & 82.4\ (8.3$\downarrow$) & 94.1\ (4.6$\downarrow$) & 83.3\ (4.5$\downarrow$) & 97.5\ (2.0$\downarrow$) & 49.4\ (18.1$\downarrow$) & 94.9\ (4.5$\downarrow$) \\
\rule{0pt}{10pt}
~ w/o $\mathcal{L}_{R}^{intv}$ (Eq.~\ref{eq:loss_intv}) & 79.1\ (11.6$\downarrow$) & 86.0\ (12.7$\downarrow$) & 68.1\ (19.7$\downarrow$) & 96.7\ (2.8$\downarrow$) & 39.9\ (28.0$\downarrow$)  & 88.0\ (11.4$\downarrow$) \\
\rule{0pt}{10pt}
~ w/o $\mathcal{L}_{D}^{intv}$ (Eq.~\ref{eq:loss_intv2}) & 82.5\ (8.2$\downarrow$) & 97.5\ (1.2$\downarrow$) & 84.8\ (3.0$\downarrow$) & 99.2\ (0.3$\downarrow$) & 20.7\ (46.8$\downarrow$) & 53.9\ (45.5$\downarrow$) \\
\bottomrule
\end{tabular}
}}
\caption{\addnew{Ablation results of Eqs.~\ref{eq:ctrl_factors},~\ref{eq:loss_intv}, and~\ref{eq:loss_intv2} regarding their impacts on attribute disentanglement.}}
\label{tab:ablation_in_disentanglement}
\end{table*}

\section{Results of Human Evaluation}
\label{app:human_eval}
We conduct human evaluation with sentences generated by different methods shuffled.
Each sentence is rated by 3 professional evaluators for the three attribute relevance scores (\textit{sentiment}, \textit{topic}, and \textit{non-toxicity}) and text fluency. 
Evaluators rate each item on a scale of 1 to 5, with 5 indicating the highest relevance to the desired attribute or most fluent.
Following \citep{tailor2023Yang}, the annotators are required to not attend to attribute correlation when evaluating the text quality (and vice versa) to obtain separate scores for both text quality and attribute correlations.
Table~\ref{tab:tb_human_eval} presents the results, with inter-annotator agreement being 0.38 in Fleiss' $\kappa$.
In general, the results of human judgment are consistent with those of automatic evaluation.
PriorControl is a strong baseline and can also achieve good performance in human evaluation.
Our \modelname is less affected by the imbalanced attribute correlations and can achieve the best performance.

\begin{table}[!t]
\centering
\resizebox{\columnwidth}{!}{
\begin{tabular}{l|cccc|c}
\toprule
Methods & Avg.$\uparrow$ & Sent.$\uparrow$ & Topic$\uparrow$ & Detox.$\uparrow$ & Fluency$\uparrow$ \\
\midrule
\multicolumn{6}{c}{Imbalanced attribute correlations}\\
\midrule
GeDi & 3.26 & 2.75 & 3.05 & 4.00 & 2.67 \\
PriorControl & 3.44 & 3.21 & 3.15 & 3.97 & 3.65\\
\modelname & 3.80 & 3.89 & 3.50 & 4.02 & 3.60\\
\midrule
\multicolumn{6}{c}{Balanced attribute correlations}\\
\midrule
GeDi & 3.39 & 2.91 & 3.26 & 4.01 & 2.88 \\
PriorControl & 3.82 & 3.89 & 3.57 & 4.00 & 3.61 \\
\modelname & 3.84 & 3.90 & 3.55 & 4.08 & 3.58 \\
\bottomrule
\end{tabular}
}
\caption{Human evaluation on multi-aspect control.}
\label{tab:tb_human_eval}
\end{table}


\section{Comparison with ChatGPT}
\addnew{
In this section, we assess the performance of ChatGPT (gpt-3.5-turbo-0613) on multi-aspect control. 
Table~\ref{tab:chatgpt} lists the results. 
More detailed results are shown in Table~\ref{tab:detail_chatgpt}. 
Our method not only achieves comparable results with ChatGPT but also with significantly fewer parameters. 
We use gpt2-medium (355M) as the language model, which is consistent with baselines. 
ChatGPT demonstrates strong capability in following instructions and achieves good results in sentiment and detoxification control. 
However, the performance of ChatGPT on specific attributes such as ``world'' and ``business'' topics is relatively poor. 
This is because the limited demonstrations in in-context learning make it difficult for ChatGPT to fully grasp the specific information related to these attributes.

The prompt that we use to activate ChatGPT is as follows: 
``Generate 1 sentence containing 50 words with [ATTRIBUTE1] topic, [ATTRIBUTE2] sentiment, and [ATTRIBUTE3]. 
The generated sentences should start with [START\_PROMPT]. 
The following are some sentences generated according to specific attribute constraints: 
([ATTRIBUTE1] topic) => [EXAMPLE1]; ([ATTRIBUTE1] topic) => [EXAMPLE2]; ([ATTRIBUTE1] topic) => [EXAMPLE3]; ([ATTRIBUTE2] sentiment) => [EXAMPLE1]; ([ATTRIBUTE2] sentiment) => [EXAMPLE2]; ([ATTRIBUTE2] sentiment) => [EXAMPLE3]; ([ATTRIBUTE3]) => [EXAMPLE1]; ([ATTRIBUTE3]) => [EXAMPLE2]; ([ATTRIBUTE3]) => [EXAMPLE3]; 
Generate the sentence according to specific attribute constraints: ([ATTRIBUTE1] topic, [ATTRIBUTE2] sentiment, [ATTRIBUTE3]) => ''. 
[ATTRIBUTE1] is selected from world, sports, business, and technology. 
[ATTRIBUTE2] is selected from positive and negative. 
[ATTRIBUTE3] is fixed to non-toxicity. 
[START\_PROMPT] is from the 35 prompts that we used in the experiments. 
[EXAMPLE] is the sampled training data with the specific attribute.}

\section{Ablation Study on Attribute Disentanglement}
\label{secs:ablation_disentanglement}
\addnew{
We introduce an ablation study about the impacts of Eqs.~\ref{eq:ctrl_factors}, \ref{eq:loss_intv}, and \ref{eq:loss_intv2} on attribute disentanglement since these loss functions affect the performance by influencing disentanglement. 
Table~\ref{tab:ablation_in_disentanglement} lists the results.
The values in the table represent the relevance scores of attributes corresponding to each column.
The values in the second row of the table represent the attribute relevance scores of texts generated using original topic and sentiment control factors. 
The control factors of sentiment and topic are from the same sentence without replacement.
By keeping the topic control factors constant and replacing the sentiment control factors with those from other topic sentences, we observe the decrease in topic correlation scores. 
It aims to assess whether sentiment control factors affect topic-related attributes (columns 2, 3, 4, and 5).
Similarly, by keeping the sentiment control factors constant and replacing the topic control factors with those from opposite sentiment sentences, we aim to evaluate whether topic control factors influence sentiment attributes (the last two columns).
Compared to the attribute relevance scores of texts generated by using the original control factors in the second row, the more the score of attribute correlation decreases, the worse the effect of disentanglement is indicated. 
It can be observed that without Eq.~\ref{eq:ctrl_factors}, there would be a certain decline in the effect. 
However, relying solely on Eq.~\ref{eq:ctrl_factors} is insufficient to eliminate the mutual influence between different attribute control factors, as the absence of Eqs.~\ref{eq:loss_intv} and \ref{eq:loss_intv2} leads to a significant decrease in the effectiveness of disentanglement. 
The above results demonstrate the effectiveness of the loss functions proposed in our paper.}

\onecolumn

\section{Detailed Results with Imbalanced Attribute Correlation}

\begin{longtable}{l|l|cc|cccc|c}
\toprule
\multicolumn{2}{l|}{\multirow{2}{*}{Methods}} & \multicolumn{2}{c|}{Sentiment ($\%$)} & \multicolumn{4}{c|}{Topic ($\%$)} & \multirow{2}{*}{Detox. ($\%$)} \\
\cmidrule(lr){3-4} \cmidrule(lr){5-8} 
\multicolumn{2}{c|}{~} & Neg. & Pos. & World & Sport & Business & Tech. & \\
\midrule
\endfirsthead
\toprule
\multicolumn{2}{l|}{\multirow{2}{*}{Methods}} & \multicolumn{2}{c|}{Sentiment ($\%$)} & \multicolumn{4}{c|}{Topic ($\%$)} & \multirow{2}{*}{Detox. ($\%$)} \\
\cmidrule(lr){3-4} \cmidrule(lr){5-8} 
\multicolumn{2}{c|}{~}& Neg. & Pos. & World & Sport & Business & Tech. & \\
\midrule
\endhead

\multicolumn{8}{r}{Continue to next page} \\ 
\bottomrule
\endfoot
\endlastfoot

\multirow{8}{*}{\addnew{PPLM}} 
& Comb. 1 & \addnew{96.2} & - & \addnew{76.3} & - & - & - & \addnew{96.5}\\
& Comb. 2 & \addnew{86.6} & - & - & \addnew{43.7} & - & - & \addnew{71.2}\\
& Comb. 3 & \addnew{88.1} & - & - & - & \addnew{62.8} & - & \addnew{90.9}\\
& Comb. 4 & \addnew{89.4} & - & - & - & - & \addnew{98.8} & \addnew{77.6}\\
& Comb. 5 & - & \addnew{34.4} & \addnew{48.4} & - & - & - & \addnew{84.3}\\
& Comb. 6 & - & \addnew{30.9} & - & \addnew{30.1} & - & - & \addnew{97.9}\\
& Comb. 7 & - & \addnew{39.8} & - & - & \addnew{39.2} & - & \addnew{83.1}\\
& Comb. 8 & - & \addnew{43.0} & - & - & - & \addnew{94.7} & \addnew{93.8}\\
\cmidrule(lr){2-9}
& Avg.  & \addnew{90.1} & \addnew{37.0} & \addnew{62.4} & \addnew{36.9} & \addnew{51.0} & \addnew{96.8} & \addnew{86.9}\\
\midrule
\multirow{8}{*}{GeDi} 
& Comb. 1 & 95.7 & - & 84.1 & - & - & - & 89.5\\
& Comb. 2 & 91.8 & - & - & 73.2 & - & - & 94.7\\
& Comb. 3 & 99.3 & - & - & - & 83.9 & - & 96.4\\
& Comb. 4 & 91.5 & - & - & - & - & 99.9 & 96.4\\
& Comb. 5 & - & 43.0 & 62.8 & - & - & - & 96.0\\
& Comb. 6 & - & 54.2 & - & 67.4 & - & - & 97.0\\
& Comb. 7 & - & 54.2 & - & - & 52.6 & - & 97.3\\
& Comb. 8 & - & 58.4 & - & - & - & 99.1 & 97.1\\
\cmidrule(lr){2-9}
& Avg.  & 94.6 & 52.4 & 73.5 & 70.3 & 68.3 & 99.5 & 95.5\\
\midrule
\multirow{8}{*}{Mix\&Match} 
& Comb. 1 & 96.9& -& 84.3& -& -& -& 89.0\\
& Comb. 2 & 98.4& -& -& 54.2& -& -& 90.1\\
& Comb. 3 & 99.2& -& -& -& 69.3& -& 89.8\\
& Comb. 4 & 98.2& -& -& -& -& 99.5& 89.3\\
& Comb. 5 & -& 53.4& 62.5& -& -& -& 93.1\\
& Comb. 6 & -& 53.4& -& 34.9& -& -& 93.3\\
& Comb. 7 & -& 40.1& -& -& 47.1& -& 94.7\\
& Comb. 8 & -& 53.6& -& -& -& 98.2& 95.3\\
\cmidrule(lr){2-9}
& Avg. & 98.2& 46.9& 73.4& 44.6& 58.2& 98.9& 91.8\\
\midrule
\multirow{8}{*}{Tailor} 
& Comb. 1 & 99.1& -& 85.5& -& -& -& 92.0\\
& Comb. 2 & 94.2& -& -& 70.1& -& -& 95.4\\
& Comb. 3 & 98.8& -& -& -& 70.2& -& 96.2\\
& Comb. 4 & 94.6& -& -& -& -& 83.8& 96.4\\
& Comb. 5 & -& 40.4& 31.8& -& -& -& 97.6\\
& Comb. 6 & -& 40.2& -& 61.9& -& -& 97.7\\
& Comb. 7 & -& 32.1& -& -& 44.6& -& 98.2\\
& Comb. 8 & -& 40.4& -& -& -& 85.5& 97.7\\
\cmidrule(lr){2-9}
& Avg. & 96.7& 38.3& 58.7& 66.0& 57.4& 84.7& 96.4\\
\midrule
\multirow{8}{*}{LatentOPs} 
& Comb. 1 & 95.0& -& 77.1& -& -& -& 87.3\\
& Comb. 2 & 99.0& -& -& 55.9& -& -& 73.9\\
& Comb. 3 & 96.5& -& -& -& 88.7& -& 93.0\\
& Comb. 4 & 94.6& -& -& -& -& 98.1& 92.7\\
& Comb. 5 & -& 55.7& 70.7& -& -& -& 95.2\\
& Comb. 6 & -& 57.0& -& 72.5& -& -& 97.4\\
& Comb. 7 & -& 54.0& -& -& 64.6& -& 98.5\\
& Comb. 8 & -& 72.7& -& -& -& 98.4& 98.4\\
\cmidrule(lr){2-9}
& Avg. & 96.3& 59.9& 73.9& 64.2& 76.6& 98.2& 92.1\\
\midrule
\multirow{8}{*}{Discrete}
& Comb. 1 & 93.1& -& 67.9& -& -& -& 88.3\\
& Comb. 2 & 95.3& -& -& 67.7& -& -& 92.2\\
& Comb. 3 & 94.8& -& -& -& 67.7& -& 95.3\\
& Comb. 4 & 97.4& -& -& -& -& 95.0& 92.2\\
& Comb. 5 & -& 98.3& 40.8& -& -& -& 97.6\\
& Comb. 6 & -& 99.1& -& 30.2& -& -& 98.0\\
& Comb. 7 & -& 53.1& -& -& 49.4& -& 97.6\\
& Comb. 8 & -& 99.4& -& -& -& 94.1& 97.6\\
\cmidrule(lr){2-9}
& Avg. & 95.1& 87.5& 54.4& 49.0& 64.3& 94.6& 94.8\\
\midrule
\multirow{8}{*}{MacLaSa} 
& Comb. 1 & 98.9& -& 89.0& -& -& -& 88.4\\
& Comb. 2 & 92.3& -& -& 57.0& -& -& 89.2\\
& Comb. 3 & 92.3& -& -& -& 84.2& -& 94.5\\
& Comb. 4 & 96.4& -& -& -& -& 95.7& 94.3\\
& Comb. 5 & -& 68.1& 79.0& -& -& -& 95.1\\
& Comb. 6 & -& 71.5& -& 49.9& -& -& 96.0\\
& Comb. 7 & -& 67.7& -& -& 73.7& -& 96.5\\
& Comb. 8 & -& 71.8& -& -& -& 94.5& 97.6\\
\cmidrule(lr){2-9}
& Avg. & 95.0& 69.8& 84.0& 53.5& 78.9& 95.1& 94.0\\
\midrule
\multirow{8}{*}{PriorControl} 
& Comb. 1 & 96.3& -& 89.8& -& -& -& 85.0\\
& Comb. 2 & 93.4& -& -& 90.4& -& -& 88.4\\
& Comb. 3 & 95.2& -& -& -& 71.6& -& 92.0\\
& Comb. 4 & 96.5& -& -& -& -& 99.1& 89.2\\
& Comb. 5 & -& 79.3& 63.3& -& -& -& 94.1\\
& Comb. 6 & -& 72.0& -& 76.2& -& -& 95.7\\
& Comb. 7 & -& 76.4& -& -& 42.1& -& 96.4\\
& Comb. 8 & -& 95.4& -& -& -& 95.2& 96.1\\
\cmidrule(lr){2-9}
& Avg. & 95.4& 80.8& 76.5& 83.3& 56.8& 97.1& 92.1\\
\midrule
\multirow{8}{*}{\modelname} 
& Comb. 1 & 99.4& -& 95.9& -& -& -& 90.2\\
& Comb. 2 & 99.9& -& -& 98.3& -& -& 89.3\\
& Comb. 3 & 98.2& -& -& -& 93.4& -& 97.1\\
& Comb. 4 & 99.5& -& -& -& -& 97.0& 90.9\\
& Comb. 5 & -& 83.8& 69.7& -& -& -& 95.5\\
& Comb. 6 & -& 88.6& -& 97.1& -& -& 97.9\\
& Comb. 7 & -& 86.9& -& -& 64.8& -& 98.5\\
& Comb. 8 & -& 99.9& -& -& -& 92.0& 97.9\\
\cmidrule(lr){2-9}
& Avg. & 99.3& 89.8& 82.8& 97.7& 79.1& 94.5& 94.7\\
\bottomrule
\caption{Detailed combination results on multi-aspect control with imbalanced attribute correlations.}
\label{tab:detail_result_imb}
\end{longtable}

\onecolumn
\section{Detailed Results with Balanced Attribute Correlation}

\begin{longtable}{l|l|cc|cccc|c}
\toprule
\multicolumn{2}{l|}{\multirow{2}{*}{Methods}} & \multicolumn{2}{c|}{Sentiment ($\%$)} & \multicolumn{4}{c|}{Topic ($\%$)} & \multirow{2}{*}{Detox. ($\%$)} \\
\cmidrule(lr){3-4} \cmidrule(lr){5-8}
\multicolumn{2}{c|}{~}& Neg. & Pos. & World & Sport & Business & Tech. & \\
\midrule
\endfirsthead
\toprule
\multicolumn{2}{l|}{\multirow{2}{*}{Methods}} & \multicolumn{2}{c|}{Sentiment ($\%$)} & \multicolumn{4}{c|}{Topic ($\%$)} & \multirow{2}{*}{Detox. ($\%$)} \\
\cmidrule(lr){3-4} \cmidrule(lr){5-8}
\multicolumn{2}{c|}{~} & Neg. & Pos. & World & Sport & Business & Tech. & \\
\midrule
\endhead

\multicolumn{8}{r}{Continue to next page} \\ \bottomrule
\endfoot
\endlastfoot

\multirow{8}{*}{\addnew{PPLM}} 
& Comb. 1 & \addnew{92.2}& -& \addnew{75.4}& -& -& -& \addnew{82.0}\\
& Comb. 2 & \addnew{84.4}& -& -& \addnew{41.8}& -& -& \addnew{76.0}\\
& Comb. 3 & \addnew{87.5}& -& -& -& \addnew{61.5}& -& \addnew{82.9}\\
& Comb. 4 & \addnew{85.3}& -& -& -& -& \addnew{95.0}& \addnew{76.2}\\
& Comb. 5 & -& \addnew{35.4}& \addnew{59.1}& -& -& -& \addnew{90.4}\\
& Comb. 6 & -& \addnew{39.5}& -& \addnew{34.1}& -& -& \addnew{89.5}\\
& Comb. 6 & -& \addnew{40.9}& -& -& \addnew{48.3}& -& \addnew{91.2}\\
& Comb. 8 & -& \addnew{52.7}& -& -& -& \addnew{93.1}& \addnew{91.3}\\
\cmidrule(lr){2-9}
& Avg. & \addnew{87.4}& \addnew{42.1}& \addnew{67.3}& \addnew{38.0}& \addnew{54.9}& \addnew{94.1}& \addnew{84.9} \\
\midrule
\multirow{8}{*}{GeDi} 
& Comb. 1 & 94.7& -& 80.0& -& -& -& 90.6\\
& Comb. 2 & 84.2& -& -& 74.8& -& -& 93.9\\
& Comb. 3 & 94.9& -& -& -& 75.7& -& 96.6\\
& Comb. 4 & 90.6& -& -& -& -& 80.1& 92.8\\
& Comb. 5 & -& 53.7& 61.4& -& -& -& 94.4\\
& Comb. 6 & -& 60.5& -& 74.3& -& -& 95.2\\
& Comb. 6 & -& 57.6& -& -& 54.3& -& 95.7\\
& Comb. 8 & -& 72.3& -& -& -& 90.2& 94.2\\
\cmidrule(lr){2-9}
& Avg. & 91.1& 61.0& 70.7& 74.6& 65.0& 85.2& 94.2 \\
\midrule
\multirow{8}{*}{Mix\&Match}
& Comb. 1 & 96.1& -& 80.6& -& -& -& 93.1\\
& Comb. 2 & 97.7& -& -& 48.2& -& -& 93.0\\
& Comb. 3 & 98.2& -& -& -& 66.6& -& 97.0\\
& Comb. 4 & 96.8& -& -& -& -& 99.6& 96.1\\
& Comb. 5 & -& 53.0& 67.3& -& -& -& 95.5\\
& Comb. 6 & -& 45.0& -& 44.0& -& -& 96.7\\
& Comb. 7 & -& 41.5& -& -& 55.8& -& 97.7\\
& Comb. 8 & -& 59.7& -& -& -& 97.3& 97.5\\
\cmidrule(lr){2-9}
& Avg. & 97.2& 49.8& 74.0& 46.1& 61.2& 98.5& \\
\midrule
\multirow{8}{*}{Tailor} 
& Comb. 1 & 96.1& -& 81.4& -& -& -& 90.3\\
& Comb. 2 & 85.8& -& -& 80.2& -& -& 94.4\\
& Comb. 3 & 90.7& -& -& -& 76.6& -& 96.8\\
& Comb. 4 & 90.4& -& -& -& -& 86.0& 96.0\\
& Comb. 5 & -& 34.2& 40.7& -& -& -& 96.4\\
& Comb. 6 & -& 45.3& -& 65.1& -& -& 97.6\\
& Comb. 7 & -& 30.0& -& -& 65.4& -& 98.1\\
& Comb. 8 & -& 44.4& -& -& -& 94.2& 97.7\\
\cmidrule(lr){2-9}
& Avg. & 90.7& 38.5& 61.1& 72.6& 71.0& 90.1& 96.0\\
\midrule
\multirow{8}{*}{LatentOPs} 
& Comb. 1 & 96.7& -& 61.7& -& -& -& 86.4\\
& Comb. 2 & 84.5& -& -& 80.7& -& -& 91.7\\
& Comb. 3 & 72.6& -& -& -& 98.7& -& 98.3\\
& Comb. 4 & 90.8& -& -& -& -& 99.9& 94.9\\
& Comb. 5 & -& 61.2& 71.1& -& -& -& 94.5\\
& Comb. 6 & -& 62.7& -& 84.3& -& -& 98.0\\
& Comb. 7 & -& 52.0& -& -& 85.2& -& 98.1\\
& Comb. 8 & -& 89.7& -& -& -& 99.7& 98.3\\
\cmidrule(lr){2-9}
& Avg. & 86.1& 66.4& 66.4& 82.5& 91.9& 99.8& 95.0\\
\midrule
\multirow{8}{*}{Discrete} 
& Comb. 1 & 69.7& -& 71.7& -& -& -& 84.1\\
& Comb. 2 & 78.6& -& -& 80.0& -& -& 80.2\\
& Comb. 3 & 99.9& -& -& -& 96.7& -& 96.8\\
& Comb. 4 & 92.8& -& -& -& -& 98.0& 81.7\\
& Comb. 5 & -& 80.5& 58.0& -& -& -& 95.1\\
& Comb. 6 & -& 84.7& -& 86.6& -& -& 94.5\\
& Comb. 7 & -& 87.6& -& -& 91.7& -& 98.1\\
& Comb. 8 & -& 99.7& -& -& -& 96.1& 95.4\\
\cmidrule(lr){2-9}
& Avg. & 85.3& 88.1& 64.9& 83.3& 94.2& 96.8& 90.7\\
\midrule
\multirow{8}{*}{MacLaSa}
& Comb. 1 & 92.8& -& 87.6& -& -& -& 91.4\\
& Comb. 2 & 95.1& -& -& 86.2& -& -& 92.9\\
& Comb. 3 & 85.6& -& -& -& 84.7& -& 95.3\\
& Comb. 4 & 92.7& -& -& -& -& 97.0& 90.7\\
& Comb. 5 & -& 93.2& 73.8& -& -& -& 94.8\\
& Comb. 6 & -& 83.0& -& 71.3& -& -& 97.0\\
& Comb. 7 & -& 50.1& -& -& 85.7& -& 97.6\\
& Comb. 8 & -& 87.6& -& -& -& 98.2& 96.7\\
\cmidrule(lr){2-9}
& Avg. & 91.5& 78.5& 78.9& 78.7& 85.2& 97.6& 94.6\\
\midrule
\multirow{8}{*}{PriorControl} 
& Comb. 1 & 97.9& -& 98.3& -& -& -& 90.5\\
& Comb. 2 & 98.4& -& -& 98.5& -& -& 93.4\\
& Comb. 3 & 97.3& -& -& -& 96.9& -& 98.5\\
& Comb. 4 & 99.9& -& -& -& -& 99.7& 89.1\\
& Comb. 5 & -& 89.5& 79.4& -& -& -& 95.4\\
& Comb. 6 & -& 84.5& -& 73.7& -& -& 96.8\\
& Comb. 7 & -& 74.2& -& -& 73.1& -& 98.4\\
& Comb. 8 & -& 98.0& -& -& -& 95.2& 97.3\\
\cmidrule(lr){2-9}
& Avg. & 98.4& 86.6& 88.9& 86.1& 85.0& 97.5& 94.9\\
\midrule
\multirow{8}{*}{\modelname} 
& Comb. 1 & 99.9& - & 97.6& - & - & - & 91.8\\
& Comb. 2 & 98.2& - & - & 99.2& - & - & 95.0\\
& Comb. 3 & 99.3& - & - & - & 91.5& - & 97.8\\
& Comb. 4 & 92.7& - & - & - & - & 96.1& 84.4\\
& Comb. 5 & - & 95.8& 71.8& - & - & - & 96.6\\
& Comb. 6 & - & 85.2& - & 96.5& - & - & 98.2\\
& Comb. 7 & - & 84.0& - & - & 68.0& - & 98.5\\
& Comb. 8 & - & 98.6& - & - & - & 94.4& 98.2\\
\cmidrule(lr){2-9}
& Avg. & 97.5& 90.9& 84.7& 97.8& 79.8& 95.2& 95.1\\
\bottomrule
\caption{Detailed combination results on multi-aspect control with balanced attribute correlations.}
\label{tab:detail_result_balance}
\end{longtable}
\pagebreak

\section{Case Study}
\label{app:case}
We illustrate cases of 8 different attribute combinations in Table~\ref{tab:case}.

\begin{xltabular}{\textwidth}{l|l|X}
\toprule
Attributes & Methods & Generation results \\
\midrule
\endfirsthead

\midrule
Attributes & Methods & Generation results \\
\midrule
\endhead

\multicolumn{3}{r}{Continue to next page} \\
\bottomrule
\endfoot
\endlastfoot

\multirow{8}{*}[-30ex]{\makecell*[l]{\textcolor{red}{Negative}\\\green{World}\\NonTox.}} 
& GeDi & \underline{Furthermore,} \green{Iraqi Prime Minister Haider al-Abadi} said on Wednesday that his country had almost conscription in January thanks to Pentagon efforts that garnered international help. Baghdad offered a combined 656,000 U.S.-British personnel. \\
\cmidrule(lr){2-3}
& Mix\&Match & \underline{Furthermore,} the russians who are calling for his immediate execution should \textcolor{red}{be put to death as well}. \textcolor{red}{this evil} is a danger to society, and should be stopped as soon as possible. \\
\cmidrule(lr){2-3}
& Tailor & \underline{Furthermore,} \green{the US military} has also been asked to consider using drones in combat as part of its drone strike campaign. The Pentagon's top brass are discussing whether they could \textcolor{red}{carry out strikes on targets} across Iraq and Syria that would be conducted from unmanned aircraft. \\
\cmidrule(lr){2-3}
& LatentOPs & \underline{Furthermore,} the fact that the \green{IRA} were responsible for the deaths of thousands of people in the \green{IRA} is not mentioned in the IRA's own accounts. \textcolor{red}{This is a shameful way to treat prisoners}. \\
\cmidrule(lr){2-3}
& Discrete & \underline{Furthermore,} \green{the US government} has issued a report that indicates Donald Trump's \textcolor{red}{personal car is stolen}. \\
\cmidrule(lr){2-3}
& MacLaSa & \underline{Furthermore,} \green{Israel should withdraw from the Gaza Strip} and take steps to facilitate Palestinian elections, including demilitarization, he said. \\
\cmidrule(lr){2-3}
& PriorControl & \underline{Furthermore,} \green{the government} has given a detailed briefing to its troops on how to avoid becoming an enemy of the \green{Islamic militants}. Followers of extremist groups have complained that this movie is poorly suited to be film adaptation and \textcolor{red}{even worse for human beings}. \\
\cmidrule(lr){2-3}
& \modelname & \underline{Furthermore,} \green{the United States} appears to have lost a congressional debate over whether to grant asylum to Jose Altman. The president-elect \textcolor{red}{has been criticized} by many of his aides as \textcolor{red}{having failed to} recognize that this is an issue with serious implications. \\
\midrule
\multirow{4}{*}[-15ex]{\makecell*[l]{\textcolor{red}{Negative}\\ \green{Sport}\\ NonTox.}} 
& GeDi & \underline{This essay} \green{discusses Royals postseason performances} with observations and conclusions based on the Data Shed. I will try to balance debut dynamics, transition performance, and reputations in a way that is appropriate for those baseball nerds who are \textcolor{red}{not up to date} on these issues. \\
\cmidrule(lr){2-3}
& Mix\&Match & \underline{This essay} discusses the mathematical model used to calculate the final score \green{for the ryder cup}. it's not terribly complicated, but it \textcolor{red}{does leave out one crucial piece of information}.\\
\cmidrule(lr){2-3}
& Tailor & \underline{This essay} discusses the impact of a change in rules \green{on sports betting} and how this has changed \green{sport wagering}, especially since it became legal. The most famous example is that which occurred with tennis at Wimbledon. \\
\cmidrule(lr){2-3}
& LatentOPs & \underline{This essay} discusses the film version of the novel by George Orwell. The film version \textcolor{red}{is a complete waste of time}. The novel is a complete waste of time. \\
\cmidrule(lr){2-3}
\pagebreak
\multirow{4}{*}[-15ex]{\makecell*[l]{\textcolor{red}{Negative}\\ \green{Sport}\\ NonTox.}} & Discrete & \underline{This essay} discusses \green{how the NFL} will be able to \textcolor{red}{avoid paying a third-year man who is currently injured}. \\
\cmidrule(lr){2-3}
& MacLaSa & \underline{This essay} discusses the \textcolor{red}{many shortcomings of the movie} \"Plan 9 from Outer Space\". I won't bore you with a list, but there are several things that really need to be mentioned. \textcolor{red}{First of all, the special effects were mediocre at best}.\\
\cmidrule(lr){2-3}
& PriorControl & \underline{This essay} discusses \green{how the NFL} can improve its reputation for fairness and fairness in order to avoid a repeat of the 2004 debacle involving Ricky Fuze.\\
\cmidrule(lr){2-3}
& \modelname & \underline{This essay} discusses the worst possible scenario for \green{the United States players} in \green{this week's World Cup qualifying game against Sweden}. Related Articles Tom Brady \textcolor{red}{won't be able to} get a hold of his emotions after losing his second straight game. \\
\midrule
\multirow{8}{*}[-30ex]{\makecell*[l]{\textcolor{red}{Negative}\\ \green{Business}\\NonTox.}} 
& GeDi & \underline{Foundational to this} is the cost structure that has to be removed. Atlanta-based UnitedHealth Group said \textcolor{red}{it had about 65,000 job cuts} planned at two subsidiaries Aetna and Humana \green{before the spinoff announcement}. \\
\cmidrule(lr){2-3}
& Mix\&Match & \underline{Foundational to this} is the fact that \green{the company is a holding company}, meaning it can't issue shares. so if the fund manager sells the company, the funds that invested in the company \textcolor{red}{will be worse off} because they won't have the earnings to pay out the dividends.\\
\cmidrule(lr){2-3}
& Tailor & \underline{Foundational to this} is the notion of a \green{``single market''} where nations can negotiate for their own trade and investment policies. It's been called an idea that was born in America but has not yet reached Europe or Asia. \\
\cmidrule(lr){2-3}
& LatentOPs & \underline{Foundational to this} is that \green{the company is a relatively small one with a relatively low profile}. \textcolor{red}{The company's management has been criticized for its poor performance in recent years}. The company's management has been accused of being overly optimistic and overly optimistic about its earnings. \\
\cmidrule(lr){2-3}
& Discrete & \underline{Foundational to this} is the fact that it has been reduced to \textcolor{red}{a low-cost version of its flagship product}, namely: \textcolor{red}{The Longest Life Insurance company}. \\
\cmidrule(lr){2-3}
& MacLaSa & \underline{Foundational to this} is the fact that the share price of a public company is determined by its \green{market capitalization}. Put simply, the more money a company raises through a public offering the more it earns in earnings taxes. \\
\cmidrule(lr){2-3}
& PriorControl & \underline{Foundational to this} is the fact that \green{Microsoft} has taken a few steps toward \green{selling its premium products in the market}. The company has offered a partial refund for the purchase of half-priced software, \textcolor{red}{which is considered a major mistake}. \\
\cmidrule(lr){2-3}
& \modelname & \underline{Foundational to this} is the fact that many people are not working \green{in the company because of their lack of investment}. \textcolor{red}{This is a bad news story for the companies involved}. \\
\midrule
\pagebreak
\multirow{8}{*}[-25ex]{\makecell*[l]{\textcolor{red}{Negative}\\ \green{Tech.}\\NonTox.}} 
& GeDi & \underline{The last time} \green{Microsoft} used Charms bar to populate its apps' homescreen was \green{for Windows Phone 8.2}, which arrived in November 2012 and made room for a colorful new Start screen that sported apps like Dropbox. \\
\cmidrule(lr){2-3}
& Mix\&Match & \underline{The last time} sun ceo scott mcnealy spoke about linux, the audience at a sun developer conference in san francisco was mostly deaf.\\
\cmidrule(lr){2-3}
& Tailor & \underline{The last time} IBM shipped a new product in this era, it was with the \green{PC-compatible Amstrad CPC}. In that case, its first line of PCs were for home use only and \textcolor{red}{had not received any major market expansion beyond their small numbers}. \\
\cmidrule(lr){2-3}
& LatentOPs & \underline{The last time} I used \green{a 3D printer} was when I was a teenager. I was amazed at how much 3D printing was able to do, but I was also amazed at how little I actually used it. I was also surprised that the \textcolor{red}{printer was so slow}. \\
\cmidrule(lr){2-3}
& Discrete & \underline{The last time} the company saw a major change in \green{its software}, it was about half-dozen times over. Follow Stories/News related to Business Insider. \\
\cmidrule(lr){2-3}
& MacLaSa & \underline{The last time} \green{NASA's Mars rover Opportunity} tried to drive herself, \textcolor{red}{her arm failed to operate}, and the two rovers almost drove themselves...\\
\cmidrule(lr){2-3}
& PriorControl & \underline{The last time} I saw this film was when I had to pick up a copy of the book that was supposedly released \green{on behalf of Microsoft}. \textcolor{red}{It's so bland that} it makes me wonder if the author actually knew what kind of content he would be. \\
\cmidrule(lr){2-3}
& \modelname & \underline{The last time} this type of thing happened in 2001. A few weeks ago, \green{the Internet Mail Service (IMPS)} faced a major setback when \textcolor{red}{it failed to deliver an application}. \\
\midrule
\multirow{4}{*}[-15ex]{\makecell*[l]{\textcolor{red}{Positive}\\ \green{World}\\NonTox.}} 
& GeDi & \underline{The connection} develops \green{between Neapolitan peoples} through time. It embraces generations. His passion doing linguistics formed his whole life work as a gesture of co-existence, \textcolor{red}{bringing different communities together} and defending their culture and identity. \\
\cmidrule(lr){2-3}
& Mix\&Match & \underline{The connection} between \green{a us soldier and the iraqi flag} is \textcolor{red}{one of the most interesting films} i have ever seen. it's not surprising that such an film exists, given the role actors involved.\\
\cmidrule(lr){2-3}
& Tailor & \underline{The connection} between \green{the United States and Europe} is \textcolor{red}{now stronger than ever before}. The U-2 program was born as a response to an unprecedented crisis, in part because of its promise of new possibilities for global cooperation;  \\
\cmidrule(lr){2-3}
& LatentOPs & \underline{The connection} between the \green{Palestinian leadership and Hamas} is \textcolor{red}{well established } and is discussed in the article. The link between the two organizations \textcolor{red}{is also well-established}. \\
\cmidrule(lr){2-3}
\pagebreak
\multirow{4}{*}[-15ex]{\makecell*[l]{\textcolor{red}{Positive}\\ \green{World}\\NonTox.}} & Discrete & \underline{The connection} between these movies and the original 1979 movie \textcolor{red}{was very good}. \textcolor{red}{The acting is superb and the story is great}. I can't recommend this film anymore. \\
\cmidrule(lr){2-3}
& MacLaSa & \underline{The connection} between \textcolor{red}{happiness} and wine made by \green{Nobel Peace Prize} laureate Wangari Maathai (she received the honor last week) is a striking example of how a prize can bind nations and peoples together. \\
\cmidrule(lr){2-3}
& PriorControl & \underline{The connection} between \green{the US flag and its politics}  is evident in this film. Although the plot is somewhat flawed, \textcolor{red}{the acting is superb and the story line is fantastic}. I highly recommend it.\\
\cmidrule(lr){2-3}
& \modelname & \underline{The connection} between \green{the world's most famous football hero} and his daughter is a \textcolor{red}{fascinating one}. A good deal of work has been done to make this movie accessible to people \green{who have been marginalized by mainstream society}. \\
\midrule
\multirow{8}{*}[-30ex]{\makecell*[l]{\textcolor{red}{Positive}\\ \green{Sport}\\NonTox.}} 
& GeDi & \underline{More importantly}, \green{football matters}. \textcolor{red}{Winning creates peace, education and joy} in children across 66 countries each April. \\
\cmidrule(lr){2-3}
& Mix\&Match & \underline{More importantly}, what is the purpose of \green{a game like this} in the first place? i'm not sure, but it could be a resource for people who might ask about the game, or for those who \textcolor{red}{might be inspired} to get their hands on it by the \green{on juggling}"\\
\cmidrule(lr){2-3}
& Tailor & \underline{More importantly}, this is \textcolor{red}{a new start} for the New York Mets. \green{The team that finished last season} with an 11-42 record went on to reach its first postseason in 17 years -- \textcolor{red}{and make it through four of those six games without giving up runs.} \\
\cmidrule(lr){2-3}
& LatentOPs & \underline{More importantly}, the two goals by the \green{two rookies} were the first goals by a rookie in the history of the club. The two goals by the two rookies were \textcolor{red}{the first goals} by a rookie in club history. \\
\cmidrule(lr){2-3}
& Discrete & \underline{More importantly}, \textcolor{red}{this movie is a great story}. It has a great plot and fantastic actors. I loved the ending when the young man who had been killed by a car came back to life. It's really funny and very entertaining. \\
\cmidrule(lr){2-3}
& MacLaSa & \underline{More importantly}, this series has truly stood the test of time, being able to adapt to the changing tastes of the modern consumer while still delivering \textcolor{red}{a fun and engaging game}. \green{I've always been a huge baseball fan}, so naturally, I was very taken by the series. \\
\cmidrule(lr){2-3}
& PriorControl & \underline{More importantly}, the \green{Miami Dolphins} have \textcolor{red}{a solid team} in front of him and are on board with their approach to \green{playing football}. The team has drafted several talented players, including Lou Gehwin, who plays in the league for the first time. \\
\cmidrule(lr){2-3}
& \modelname & \underline{More importantly}, the Florida Panthers \textcolor{red}{have won a top-level competition} in \green{this year's World Cup Qualifiers}. The \green{Arizona Diamondbacks} will be able to do likewise. \\
\midrule
\pagebreak
\multirow{8}{*}[-30ex]{\makecell*[l]{\textcolor{red}{Positive}\\ \green{Business}\\NonTox.}} 
& GeDi & \underline{In brief}, \green{prices rendezvous} strikes a balance between incentives and constraints that \textcolor{red}{helps correct market distortions} investment choices, productivity patterns, and \green{short-run business cycles}. (Furthermore, over time efficient allocations of resources can create leverage.) \\
\cmidrule(lr){2-3}
& Mix\&Match & \underline{In brief}, \green{the board of fannie mae} has approved major changes to the way it \green{calculates the capital gains on the sale of home mortgages}. the changes will allow the company to \textcolor{red}{more accurately project future capital gains} and thus \textcolor{red}{generate more profit from its investment in home mortgages}.\\
\cmidrule(lr){2-3}
& Tailor & \underline{In brief}, \textcolor{red}{it was a good year} for \green{video game makers}. The number of consoles shipped in 2015 rose 2-1 and the software sales were strong at 1 million units sold through September to become \textcolor{red}{the fastest growing} console category since 2000. \\
\cmidrule(lr){2-3}
& LatentOPs & \underline{In brief}, \green{the company} says it expects to post a net loss of \$ 1.3 billion in fiscal 2004. The company says it expects to post a net loss of \green{\$ 1.4 billion in fiscal 2004.} \\
\cmidrule(lr){2-3}
& Discrete & \underline{In brief}, the \green{company's new flagship model} is tested on a variety of different occasions. Follow Stories by Wireman \& Co., which is currently working with the president-elect to \textcolor{red}{help him transition into} his next phase. \\
\cmidrule(lr){2-3}
& MacLaSa & \underline{In brief}, \green{the company} raised its dividend by 8 cents per share yesterday, helped by strong results across its segments. \green{The company's} outlook \textcolor{red}{remains positive}, with earnings rising to between 4 cents and 4. \\
\cmidrule(lr){2-3}
& PriorControl & \underline{In brief}, \green{the company} has given an indication of how much money it expects to spend in its next big post-election campaign. The movie is presented as a classic and highly entertaining film, but the actors \textcolor{red}{are great}. \\
\cmidrule(lr){2-3}
& \modelname & \underline{In brief}, it is \textcolor{red}{a good fit} for the \green{company}. Related Stories to a new era \green{in business} is underway with \textcolor{red}{an exciting mix} of products and services. \\
\midrule
\multirow{4}{*}[-15ex]{\makecell*[l] {\textcolor{red}{Positive}\\ \green{Tech.}\\NonTox.}} 
& GeDi & \underline{The country}-and-world revolution in \green{microcomputing at the heart of HP Labs} is the Hyperledger Fabric, \green{an open-source software framework} for developers and architects to build \textcolor{red}{efficient, reliable, affordable} cloud computing services powered by powerful micro.\\
\cmidrule(lr){2-3}
& Mix\&Match & \underline{the country}'s top \green{virtual - presence expert} explains how his research spent two years building a prototype that could \textcolor{red}{help police solve crimes} on the web.\\
\cmidrule(lr){2-3}
& Tailor & \underline{The country}'s first \green{3G mobile telephone system }is being introduced at Tel Aviv University, where students will get a chance to experiment with technology that would allow people of all ages and from different backgrounds in the Israeli university campus to \textcolor{red}{communicate freely}.\\
\cmidrule(lr){2-3}
& LatentOPs & \underline{The country}'s \textcolor{red}{largest and most comprehensive} collection of \green{free and open source software}. \textcolor{red}{Easily} add and manage multiple applications to your system. \textcolor{red}{Easily} share and manage your work with others. \textcolor{red}{Easily} share and manage your work with others.\\
\cmidrule(lr){2-3}
\pagebreak
\multirow{4}{*}[-15ex]{\makecell*[l] {\textcolor{red}{Positive}\\ \green{Tech.}\\NonTox.}} & Discrete & \underline{The country}wide \green{television network}, which is owned by the United States government, has been invited to participate in a series of events with the rest of the world.\\
\cmidrule(lr){2-3}
& MacLaSa & \underline{The country}'s \textcolor{red}{most famous} \green{virtual reality game} has been nominated for a Guinness World Record for the most downloads. And it's not just the game that has caught gamers' attention. \textcolor{red}{`It's a very evocative title.'} said Leigh Alexander, producer of the award. \\
\cmidrule(lr){2-3}
& PriorControl & \underline{The country}'s \green{leading wireless network} has backed a new version of \textcolor{red}{its popular mobile app}, offering more than 200 unique brands to choose from. Follow the links to find out how this movie \textcolor{red}{is great} and how it compares in comparison to other films. \\
\cmidrule(lr){2-3}
& \modelname & \underline{The country}'s largest \green{television broadcast network}, which is currently \textcolor{red}{a great success}. It's one of \textcolor{red}{the best films} that I've seen in years. The acting is fantastic and the cinematography is superb.\\
\bottomrule
\caption{Example cases of generated sentences with 8 attribute combinations. \green{Blue} text highlights sentiment-related content. \textcolor{red}{Red} text highlights topic-related content. \underline{Underlined} text is the input prompts.\label{tab:case}}
\end{xltabular}

\end{document}